\documentclass[runningheads]{llncs}

\usepackage[final,year=2026]{eccv}

\usepackage{eccvabbrv}

\usepackage{graphicx}
\usepackage{booktabs}

\usepackage[accsupp]{axessibility}  

\usepackage{subcaption}
\usepackage{graphicx}
\usepackage{amsmath} 
\usepackage{multirow}
\usepackage{adjustbox}
\usepackage{booktabs}
\usepackage{amssymb}
\usepackage{wrapfig}
\usepackage{caption} 

\usepackage{hyperref}

\usepackage{orcidlink}

\begin{document}

\title{Deep Attention Reweighting: Post-Hoc Attention-Based Feature Aggregation in CNNs for Disentangling Core and Spurious Features under Spurious Correlations} 
\titlerunning{Deep Attention Reweighting}

\author{Chew Kin Whye\inst{1}\orcidlink{0000-1111-2222-3333} \and
Jingxian Wang\inst{1}\orcidlink{1111-2222-3333-4444}}
\authorrunning{Chew and Wang}


\institute{National University of Singapore, Singapore 117583, Singapore\
\email{\{kinwhye,wang\}@nus.edu.sg}}

\maketitle

\begin{abstract}
    Convolutional Neural Networks (CNNs) often exploit spurious correlations in datasets, learning superficially predictive yet causally irrelevant features, leading to poor generalization and fairness issues. Deep Feature Reweighting (DFR) is a post-hoc technique that reduces a trained model's reliance on spurious correlations by retraining its classification head on a target dataset. However, we show that DFR is fundamentally constrained by operating on entangled features, limiting its ability to amplify the core features while simultaneously suppressing the spurious ones. We trace this entanglement to the ubiquitous Global Average Pooling (GAP) layer, which indiscriminately collapses spatially distinct core and spurious features into a single representation. To address this, we propose Deep Attention Reweighting (DAR), a post-hoc attention-based aggregation module that replaces GAP and is retrained jointly with the classification head. DAR computes an adaptive weighting of spatial locations across feature maps, enabling selective suppression of spurious features before the collapse into entangled features. Across various datasets, metrics, and ablations, DAR consistently outperforms DFR, demonstrating that our attention-based aggregation mitigates GAP-induced entanglement and reduces spurious reliance. 
\end{abstract}
\section{Introduction}

\textit{Convolutional Neural Networks} (CNNs) have achieved strong in-distribution performance for vision tasks but often fail to generalize in the presence of \textit{spurious correlations}---features in the training data that happen to be statistically correlated with the labels but unrelated to the underlying target function. Under \textit{Empirical Risk Minimization (ERM)}, CNNs will indiscriminately exploit these shortcuts to minimize the empirical risk~\cite{arjovsky2020invariant, https://doi.org/10.48550/arxiv.2010.15775, NEURIPS2020_6cfe0e61}. However, these \textit{spurious features} do not generalize consistently, resulting in a \emph{severe lack of robustness, an inability to generalize, and unfairness}~\cite{DBLP:journals/corr/abs-2004-07780}. For instance, in the Waterbirds dataset~\cite{welinder2010cub200, Sagawa2020Distributionally}, the label (waterbird vs.\ landbird) is spuriously correlated with the background (water vs.\ land), resulting in models that rely on the superficial and non-causal background shortcuts and therefore trivially fail on counterexamples such as waterbirds on land or landbirds on water. In contrast, the \textit{core features} are the task-relevant signals that are causally predictive of the label and generalize under distribution shifts.

\textit{Deep Feature Reweighting (DFR)}~\cite{https://doi.org/10.48550/arxiv.2204.02937} is a simple yet effective post-hoc technique that retrains the last layer of an ERM-trained model on a small \textit{balanced target dataset}, defined as one where there is a uniform distribution over classes within each spurious attribute such that the spurious attribute is no longer predictive of the label. This yields strong empirical results because DFR reduces the model's reliance on spurious features by adjusting the feature importance during retraining, while increasing reliance on the strong core features learned in the feature extractor. However, we find its effectiveness limited by the \emph{entanglement between the core and spurious features}, where a single output feature encodes information from both. This entanglement forces DFR to navigate a trade-off: retaining a feature to preserve core information while including spurious correlations, or suppressing it to remove spurious influences at the cost of core information. As a result, DFR struggles to selectively suppress spurious features without inadvertently affecting core information. We further trace this entanglement largely to the \textit{Global Average Pooling} (GAP) layer~\cite{lin2014nin}, a widely used feature aggregation method in many popular CNN architectures~\cite{szegedy2015inception, he2015deep, huang2018denselyconnectedconvolutionalnetworks, sandler2019mobilenetv2invertedresidualslinear, tan2020efficientnetrethinkingmodelscaling} that uniformly averages the spatially distinct core and spurious features across each feature map into an entangled representation.

\begin{figure}[tb]
    \centering
    \includegraphics[width=0.8\textwidth]{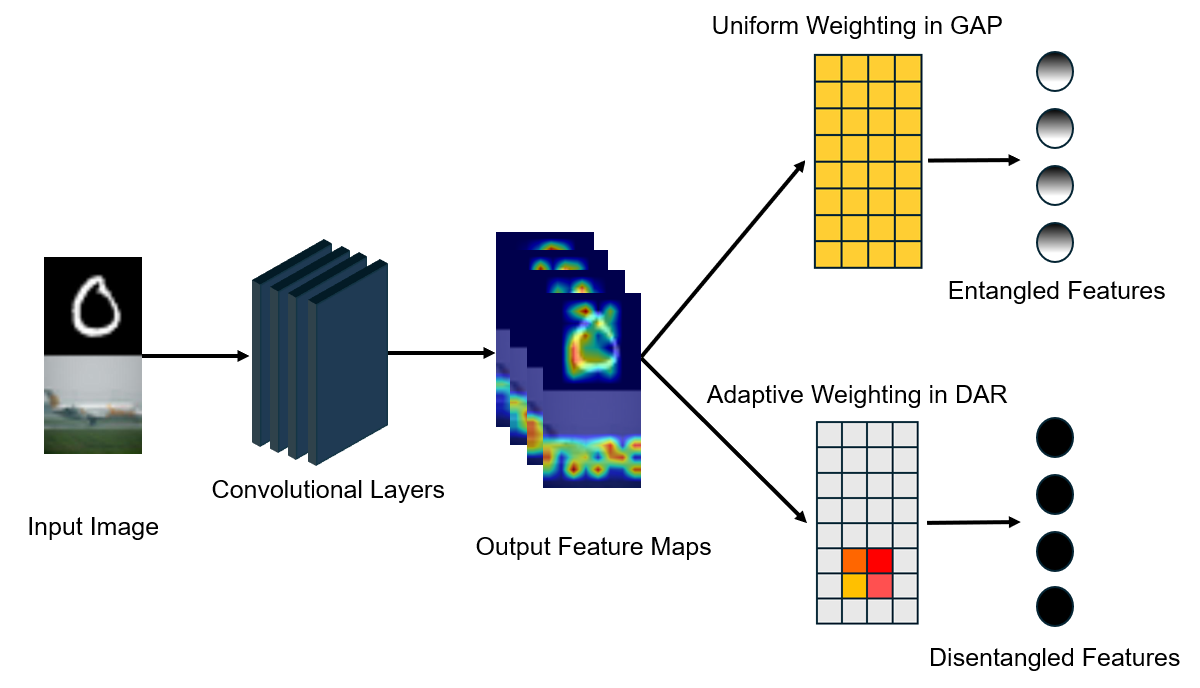}
    \caption{\textbf{Illustration of GAP vs.\ DAR.} The input image from the Dominoes dataset consists of the spurious MNIST image concatenated with the core CIFAR image. After feature extraction by the convolutional layers, we find that the output feature maps are entangled, with each feature map activating both core and spurious features at distinct spatial locations. GAP uniformly averages these feature maps across spatial locations to produce entangled output features, whereas DAR's attention mechanism computes the adaptive spatial weighting to extract a disentangled core feature representation.}
    \label{fig:gap_vs_dar}
\end{figure}

To overcome this limitation, we propose \textit{Deep Attention Reweighting} (DAR), a novel approach that replaces the GAP layer with an attention-based aggregation mechanism retrained post-hoc alongside the last layer. Whereas GAP uniformly averages all spatial locations, DAR learns an attention mechanism that \emph{adaptively identifies and extracts the spatial locations in the feature maps that correspond to core features}. This allows DAR to perform targeted feature selection at the feature map representation, where the core and spurious features are spatially disentangled prior to collapsing into the feature vector. This difference between GAP and DAR is illustrated in Figure~\ref{fig:gap_vs_dar}. 

The key contributions of this work are as follows:

\begin{itemize}
    \item \textbf{Diagnosing GAP-induced feature entanglement as the bottleneck of DFR.}
    
    We introduce the Core Effect Percentage (CEP), Core Activation Percentage (CAP), and Core GradCAM Percentage (CGP) as quantitative measures of feature entanglement, and complement them with GradCAM visualizations as a qualitative diagnostic. Using these metrics, we experimentally show that convolutional feature maps are channel-entangled yet spatially separable: core and spurious features activate within the same feature map but at different locations. However, the GAP layer collapses this spatial separability into an entangled feature vector. Consequently, DFR cannot reliably retain core information while removing spurious features.
        
    \item \textbf{Attention-based feature aggregation for disentanglement.}
    
    We propose DAR, a novel, lightweight attention-based feature aggregation mechanism that assigns adaptive importance weights to spatial locations over the convolutional feature maps. We experimentally show that when trained post-hoc on a target dataset, DAR significantly mitigates feature entanglement via the targeted extraction of core features and suppression of spurious features before they are collapsed into the final representation.
        
    \item \textbf{Comprehensive empirical validation.}

    We evaluate DAR on six spurious correlation benchmark datasets against competitive baselines and across diverse experimental configurations. Our results show that DAR consistently outperforms the baselines, demonstrating its robustness in mitigating spurious correlations. We additionally provide various targeted ablation studies that characterize the source of DAR’s performance gains.

\end{itemize}

\section{Related Works}
\label{section:related}
\subsection{Spurious Correlation}
The challenge of spurious correlation in machine learning is studied under various frameworks, including the \textit{correlation-versus-causation dilemma}~\cite{lake2016building, lopezpaz2016dependence, marcus2018deep}, \textit{invariant learning}~\cite{peters2015causal, heinzedeml2018invariant, arjovsky2020invariant}, \textit{subpopulation shift}~\cite{chen2023confidencebased, pmlr-v202-liang23d, yang2023change}, and \textit{group robustness}~\cite{Sagawa2020Distributionally, NEURIPS2023_265bee74, pmlr-v202-qiu23c}, among others (see also~\cite{ye2024spurious} for a recent survey). The methods developed to address these challenges can be broadly categorized based on their dataset requirements. Firstly, methods that rely solely on the training dataset without additional information~\cite{nam2020learning, DBLP:journals/corr/abs-2107-09044, pagliardini2022agree, taghanaki2022masktune, https://doi.org/10.48550/arxiv.2203.01517, lee2023diversify}. Secondly, methods that require datasets from different environments~\cite{pmlr-v119-ahuja20a, arjovsky2020invariant,  Lin_2022_CVPR, pmlr-v162-zhou22e}. Thirdly, methods that require \textit{group labels}~\cite{Sagawa2020Distributionally}, where each \textit{group} is defined as a combination of the label and the spurious attribute (e.g., background in Waterbirds). Lastly, methods that require a small \textit{target dataset} where spurious correlations are absent. Deep Feature Reweighting (DFR)~\cite{https://doi.org/10.48550/arxiv.2204.02937} falls under this category, retraining the final classification head of an ERM-trained model in a post-hoc manner on the target dataset, achieving state-of-the-art performance on several benchmarks.

Our work improves the applicability of DFR to CNNs by showing that its effectiveness is limited by feature entanglement in the GAP layer. We propose an attention-based aggregation mechanism that replaces it and is retrained alongside the classification head post-hoc, allowing the extraction of core features at the feature map level before they are entangled with spurious features.

\subsection{Disentangled Representation Learning}
\textit{Disentangled Representation Learning (DRL)} aims to learn a feature representation where each feature independently captures a distinct factor of variation while being invariant to the others~\cite{bengio2014representationlearningreviewnew, locatello2020disentanglingfactorsvariationusing}. A recent survey by~\cite{10579040} highlights that Variational Autoencoders~\cite{kingma2022autoencodingvariationalbayes} remain one of the most widely used frameworks, with many variants~\cite{higgins2017beta, higgins2018scanlearninghierarchicalcompositional, dupont2018learningdisentangledjointcontinuous, burgess2018understandingdisentanglingbetavae, kumar2018variationalinferencedisentangledlatent, kim2019disentanglingfactorising, kim2019relevancefactorvaelearning, mathieu2019disentanglingdisentanglementvariationalautoencoders, chen2019isolatingsourcesdisentanglementvariational} formulated under the assumption that the factors of variation are statistically independent.
However,~\cite{träuble2021disentangledrepresentationslearnedcorrelated} found that when this assumption is violated, these methods fail to prevent entanglement and may even introduce a bias against disentanglement. This makes feature disentanglement particularly challenging in the presence of spurious correlations, since the core and spurious features are statistically correlated.

Unlike DRL methods that seek to learn a disentangled representation, our method operates in a post-hoc setting where the model has already learned an entangled feature representation. Our paper identifies that the GAP layer is primarily responsible for this entanglement in CNNs and replaces it with an attention mechanism that enables targeted extraction of task-relevant features.

\subsection{Attention-Based Architectures in Vision}
According to the survey by~\cite{9716741}, the attention-based architectures in vision can be broadly categorized into two groups: pure attention architectures that replace convolutions entirely~\cite{dosovitskiy2021imageworth16x16words, touvron2021trainingdataefficientimagetransformers, liu2021swintransformerhierarchicalvision, wang2021pyramidvisiontransformerversatile, yuan2021tokenstotokenvittrainingvision} and hybrid architectures that integrate attention with convolutions~\cite{wang2018nonlocalneuralnetworks,woo2018cbamconvolutionalblockattention, hu2019squeezeandexcitationnetworks, bello2020attentionaugmentedconvolutionalnetworks, wu2020visualtransformerstokenbasedimage}. These architectures have weaker built-in spatial inductive biases and therefore often rely on large-scale pre-training to learn such structure implicitly, but are able to overcome the limited receptive field of standard convolutions. Several works have further explored the robustness of these attention architectures to spurious correlations, finding that while they can focus on semantically meaningful regions, they remain susceptible to biases~\cite{ghosal2022visiontransformersrobustspurious, wang2021causalattentionunbiasedvisual,
yang2021causalattentionvisionlanguagetasks,
10604648}.

In contrast to prior work that embeds attention within the backbone architecture to model token interactions for feature representation learning, DAR retains the CNN backbone, thereby preserving its efficiency and inductive biases. Here, attention is used as a lightweight, modular mechanism specifically designed for adaptive spatial reweighting during feature aggregation, replacing GAP.

The closest work to ours is~\cite{jetley2018learnpayattention}, which applies the simple dot-product attention to aggregate features across multiple layers. By contrast, our proposed architecture, designed for post-hoc adaptation, applies a more expressive multi-head, multi-layer attention mechanism after the final convolutional layer, where high-level, task-relevant features reside. Further implementation details and empirical comparisons are provided in Sections~\ref{section:design} and~\ref{sec:ablation}.

\section{Feature Entanglement Limits Post-Hoc Reweighting}
\label{section:entanglement}
In this section, we examine the key limitation of Deep Feature Reweighting (DFR): its reliance on entangled ERM-learned features. Leveraging our Core Effect Percentage (CEP) metric to quantify feature entanglement, our analysis reveals the following: First, the output feature maps of convolutional layers are channel-entangled but spatially disentangled. Second, uniform averaging across spatial locations via Global Average Pooling (GAP) collapses spatial separability and yields entangled output features. Finally, DFR operates on these entangled features and cannot eliminate reliance on spurious ones.

\subsection{Quantifying Feature Entanglement with Core Effect Percentage}
To quantify feature entanglement, we introduce the \textit{Core Effect Percentage} (CEP) metric, which measures the extent to which the core and spurious portions of the image influence the model’s output features and predictions. This metric is derived from the \textit{Controlled Direct Effect} in do-calculus \cite{pearl2012docalculusrevisited}, similar to many interventional-based metrics~\cite{9947342}. Each input $\mathbf{x}$ is decomposed into the core $\mathbf{x}_{core}$ and the spurious $\mathbf{x}_{spu}$ components. The effect $E$ of each component on the model $f$ is quantified by the change in the model's output when that component is independently intervened on:

\begin{align}
    \label{eqn:effect}
E_{\text{core}}(\mathbf{x}) &= \| f(\mathbf{x}) - f\!\left(\mathbf{x} \mid do(\mathbf{x}_{\text{core}} = \mathbf{x}'_{\text{core}})\right) \|_1,\\
E_{\text{spu}}(\mathbf{x}) &= \| f(\mathbf{x}) - f\!\left(\mathbf{x} \mid do(\mathbf{x}_{\text{spu}} = \mathbf{x}'_{\text{spu}})\right) \|_1.
\end{align}

Here, $\| \cdot \|_1$ denotes the L1 norm, defined as the sum of absolute values over all entries, while $\mathbf{x}'_{\text{core}}$ and $\mathbf{x}'_{\text{spu}}$ represent \textit{counterfactual replacements} of the corresponding components, instantiated by substituting that component with the value from another datapoint while keeping the other component fixed. Finally, the CEP value is the expectation over inputs of the normalized core effect:

\begin{equation}
    \label{eqn:csp}
\text{CEP} = \mathbb{E}_{\mathbf{x}}\!\left[
\frac{E_{\text{core}}(\mathbf{x})}{E_{\text{core}}(\mathbf{x}) + E_{\text{spu}}(\mathbf{x})}
\right] \times 100\% .
\end{equation}
High CEP ($\approx100\%$) indicates reliance on core features; low CEP ($\approx 0\%$) indicates reliance on spurious features; intermediate CEP ($\approx 50\%$) signifies reliance on both, i.e., entanglement.

\subsection{Empirical Analysis of Feature Entanglement}
\label{section:feresults}
\begin{figure}[tb]
  \hspace*{\fill}   
  \begin{subfigure}{0.23\textwidth}
  \includegraphics[width=\linewidth]{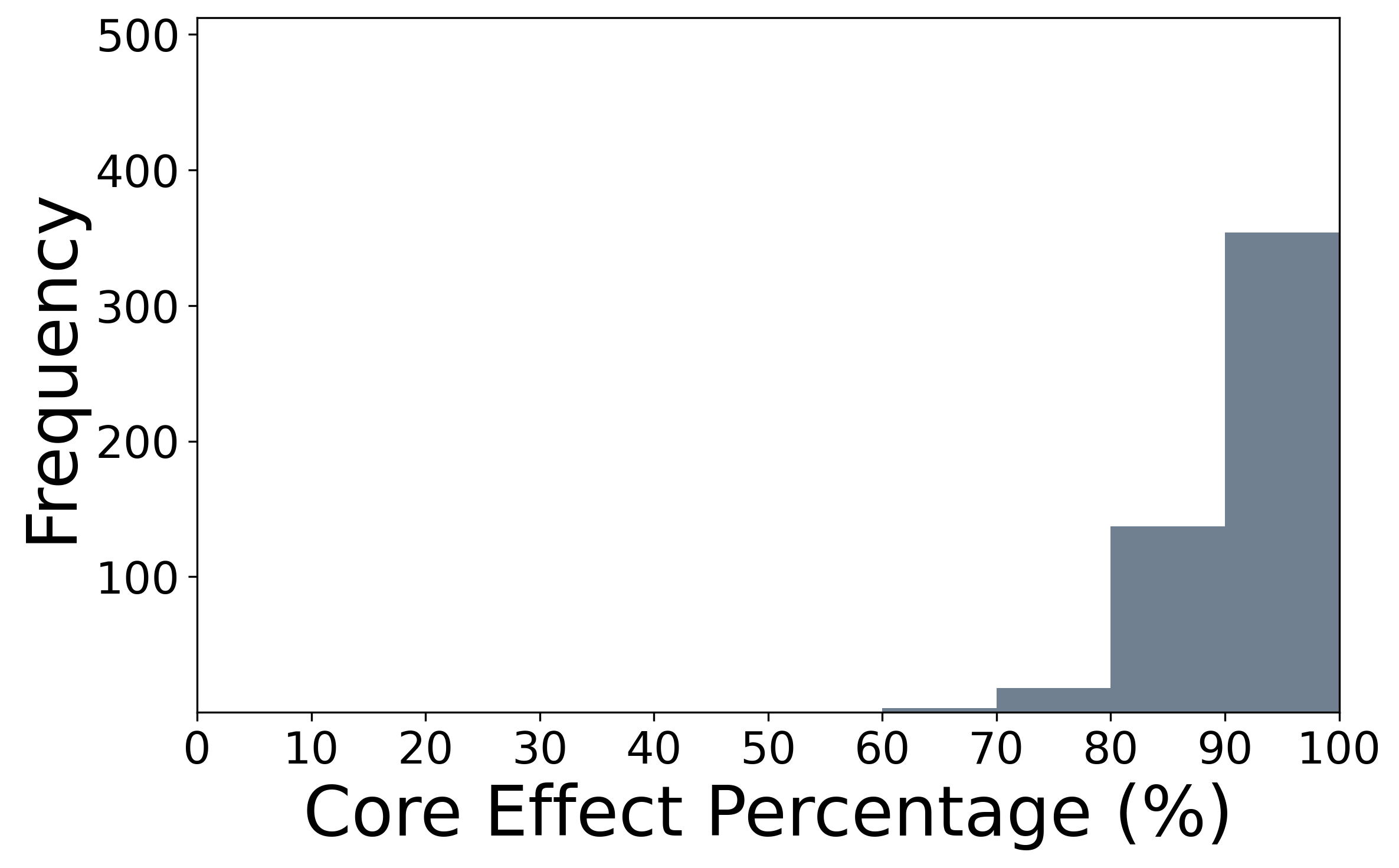}\hfill
  \caption{\(ERM_{Core}\ Map\)}
  \label{fig:map_cep_erm_core}
  \end{subfigure}%
  \hspace*{\fill}   
    \begin{subfigure}{0.23\textwidth}
  \includegraphics[width=\linewidth]{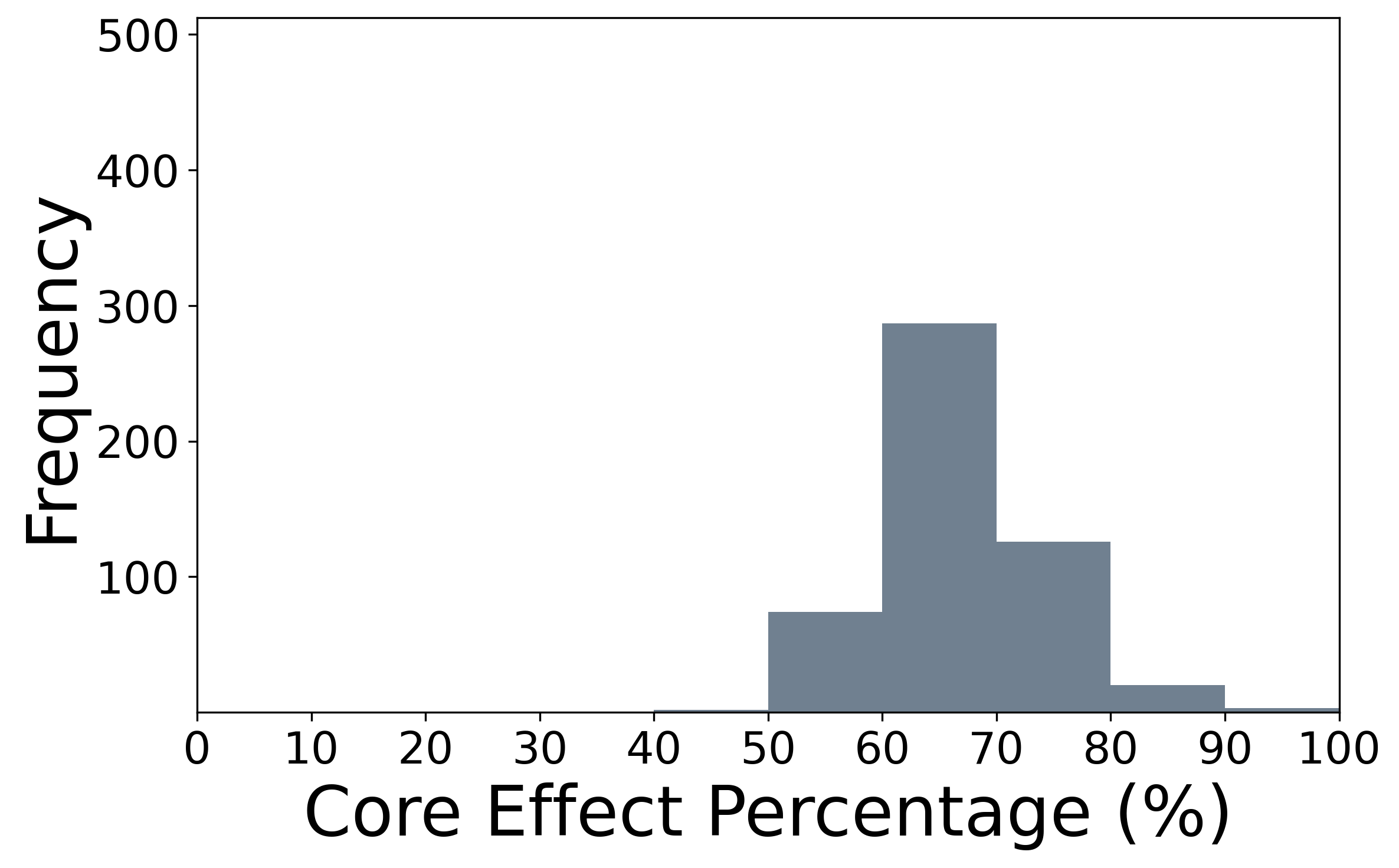}\hfill
  \caption{\(ERM\ Map\)}
  \label{fig:map_cep_erm}
  \end{subfigure}%
    \hspace*{\fill}   
  \begin{subfigure}{0.23\textwidth}
  \includegraphics[width=\linewidth]{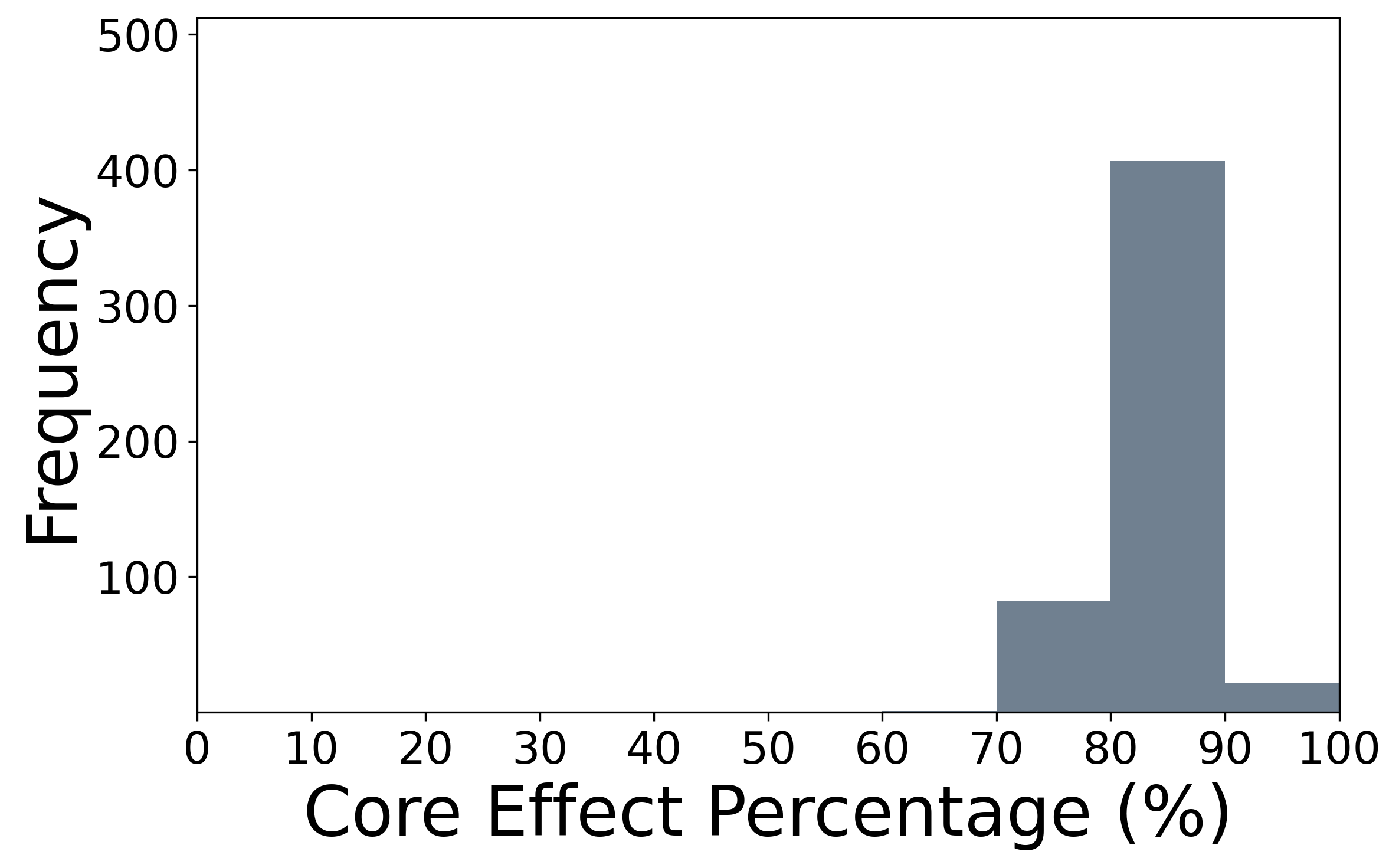}
  \caption{\(DFR_{CNN}\ Map\)} 
  \label{fig:map_cep_cnn}
  \end{subfigure}
    \hspace*{\fill}   
  \begin{subfigure}{0.23\textwidth}
  \includegraphics[width=\linewidth]{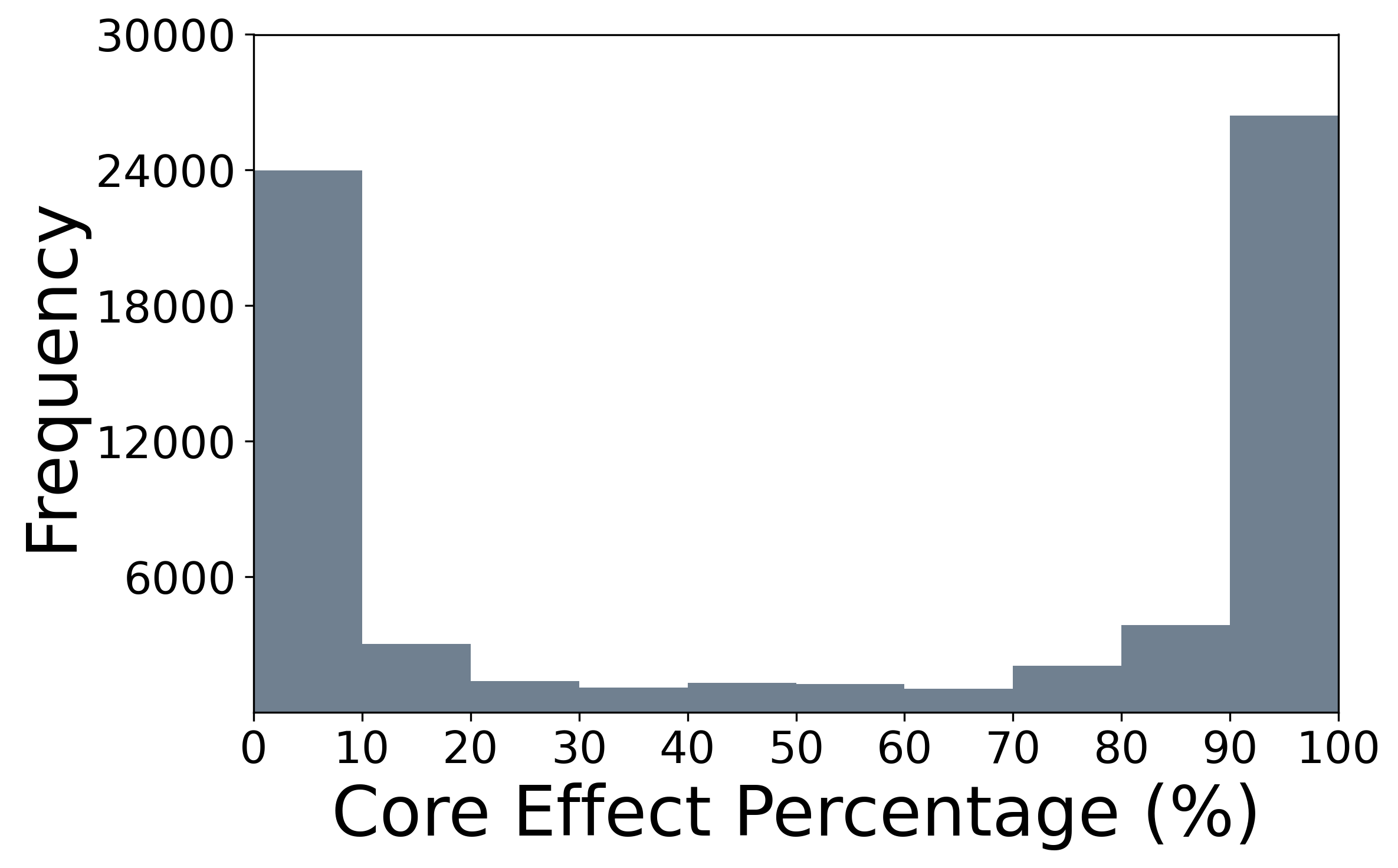}\hfill
  \caption{\(ERM\ Pixel\)} 
  \label{fig:pixel_cep_erm}
  \end{subfigure}
\hspace*{\fill}   

\caption{Histogram of CEP values across 512 output feature maps for various methods. Figures (a), (b), and (c) analyze the feature maps as a whole, whereas Figure (d) analyzes the feature maps at the pixel level. Refer to Section~\ref{section:feresults} for a detailed analysis.} \label{fig:map_cep}
\end{figure}

\begin{figure}[tb]
  \hspace*{\fill}   
  \begin{subfigure}{0.18\textwidth}
    \includegraphics[width=\linewidth]{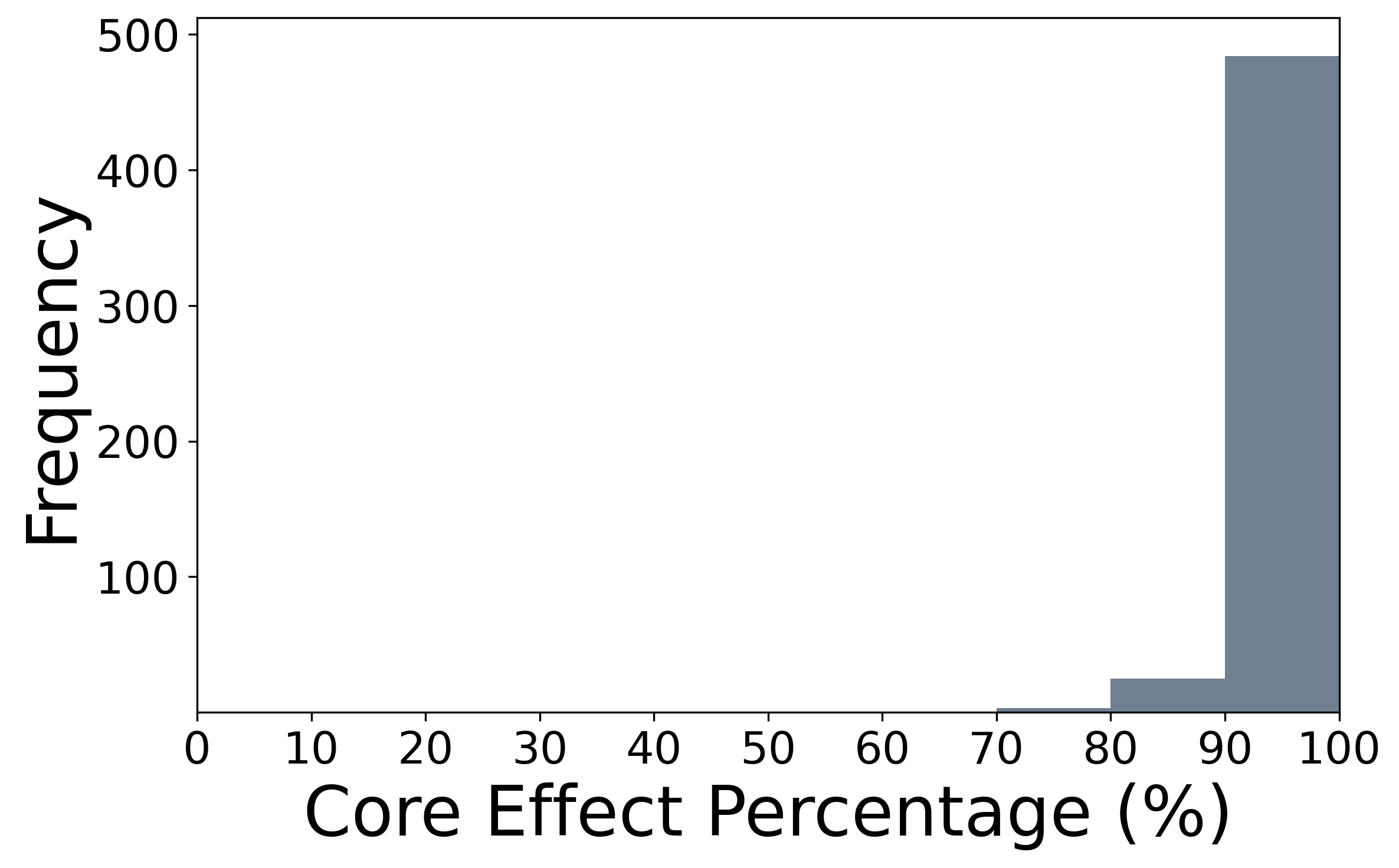}\hfill
    \caption{\(ERM_{Core}\)} 
    \label{fig:f_cep_erm_core}
  \end{subfigure}%
  \hspace*{\fill}   
  \begin{subfigure}{0.18\textwidth}
  \includegraphics[width=\linewidth]{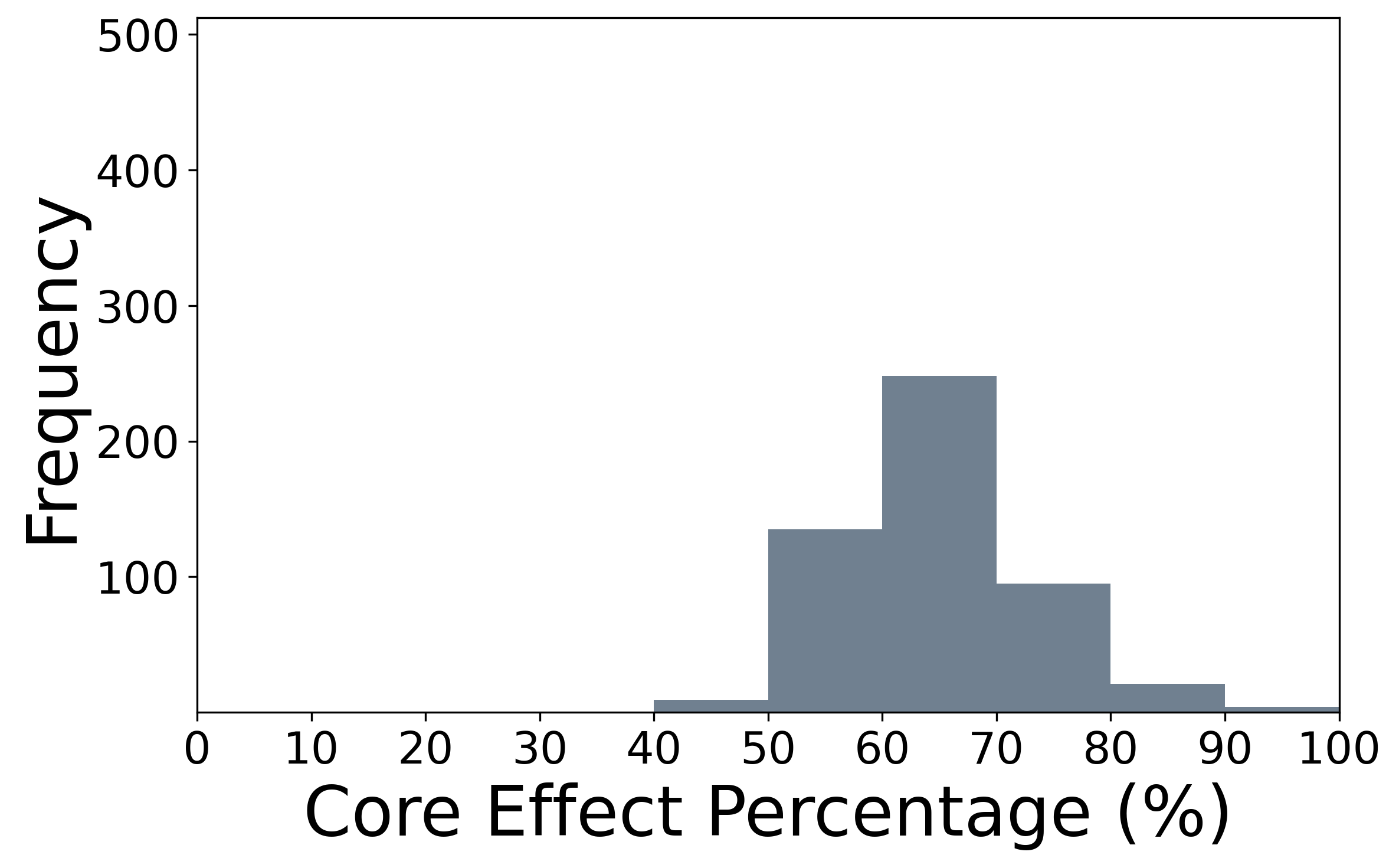}\hfill
  \caption{\(ERM\)} 
  \label{fig:f_cep_erm}
  \end{subfigure}
    \hspace*{\fill}   
  \begin{subfigure}{0.18\textwidth}
  \includegraphics[width=\linewidth]{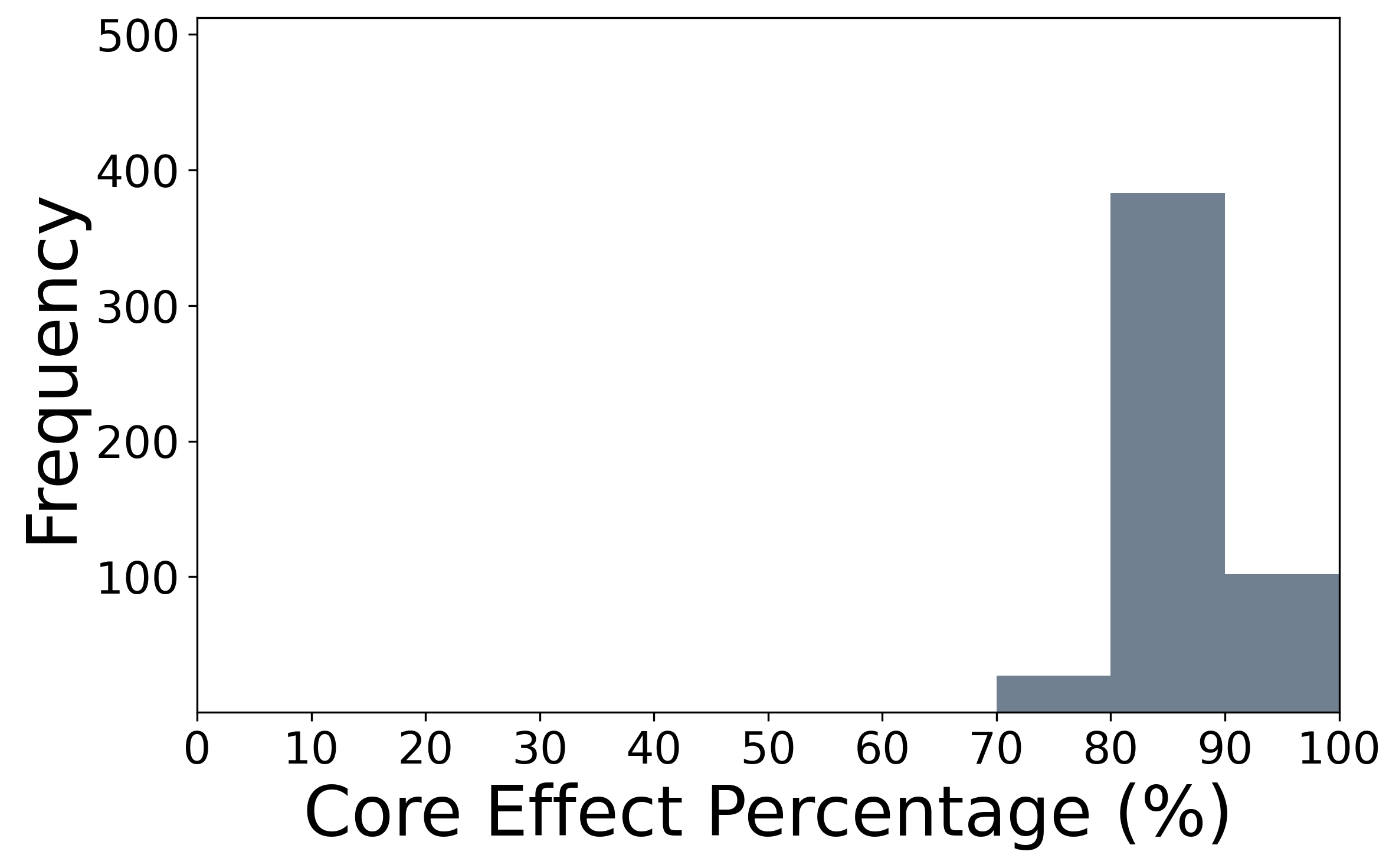}\hfill
  \caption{\(DFR_{FC}\)}
  \label{fig:f_cep_fc}
  \end{subfigure}
  \hspace*{\fill}   
    \begin{subfigure}{0.18\textwidth}
  \includegraphics[width=\linewidth]{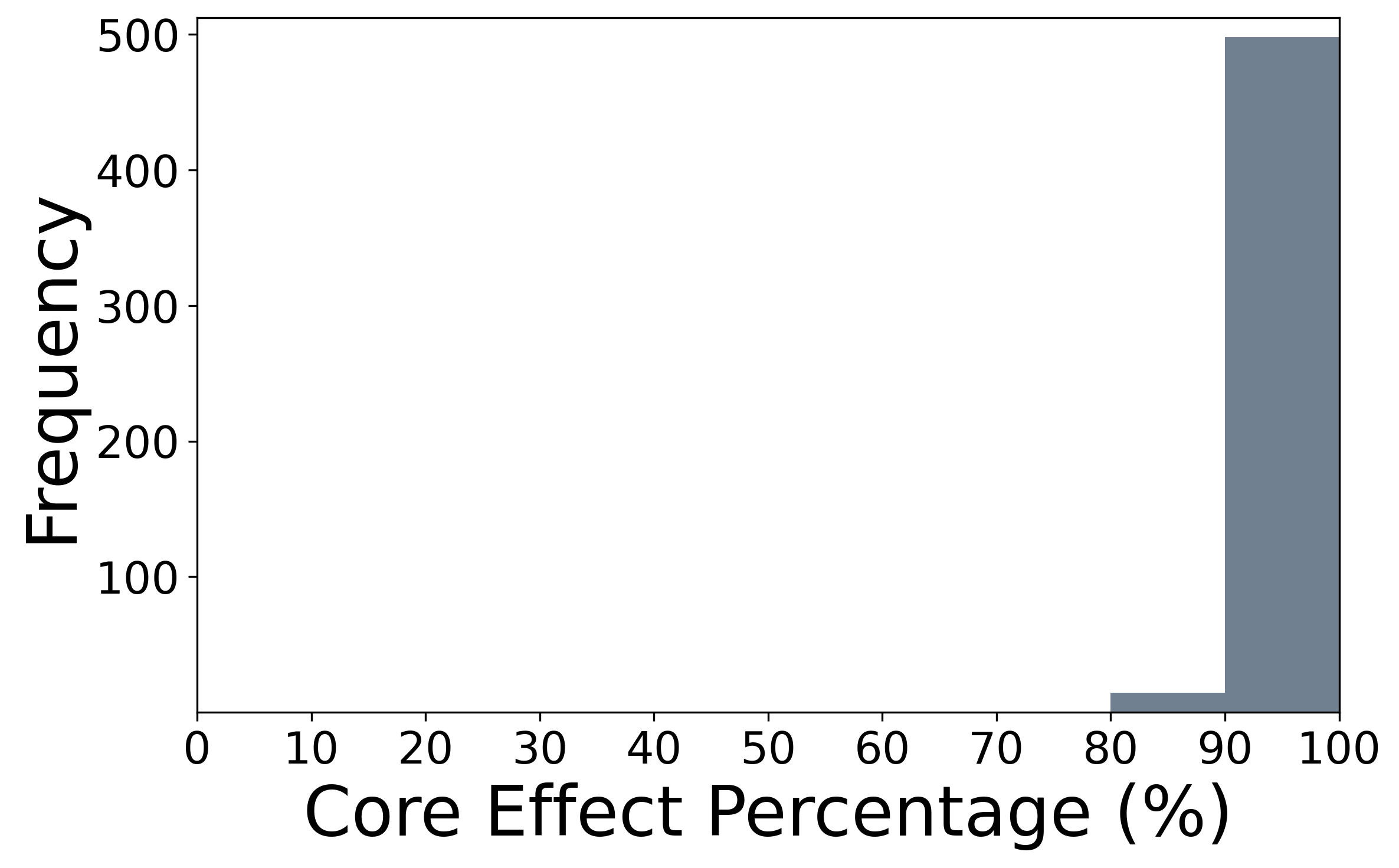}\hfill
  \caption{\(DAR\)}
  \label{fig:f_cep_dar}
  \end{subfigure}
  \hspace*{\fill}   
  \begin{subfigure}{0.18\textwidth}
  \includegraphics[width=\linewidth]{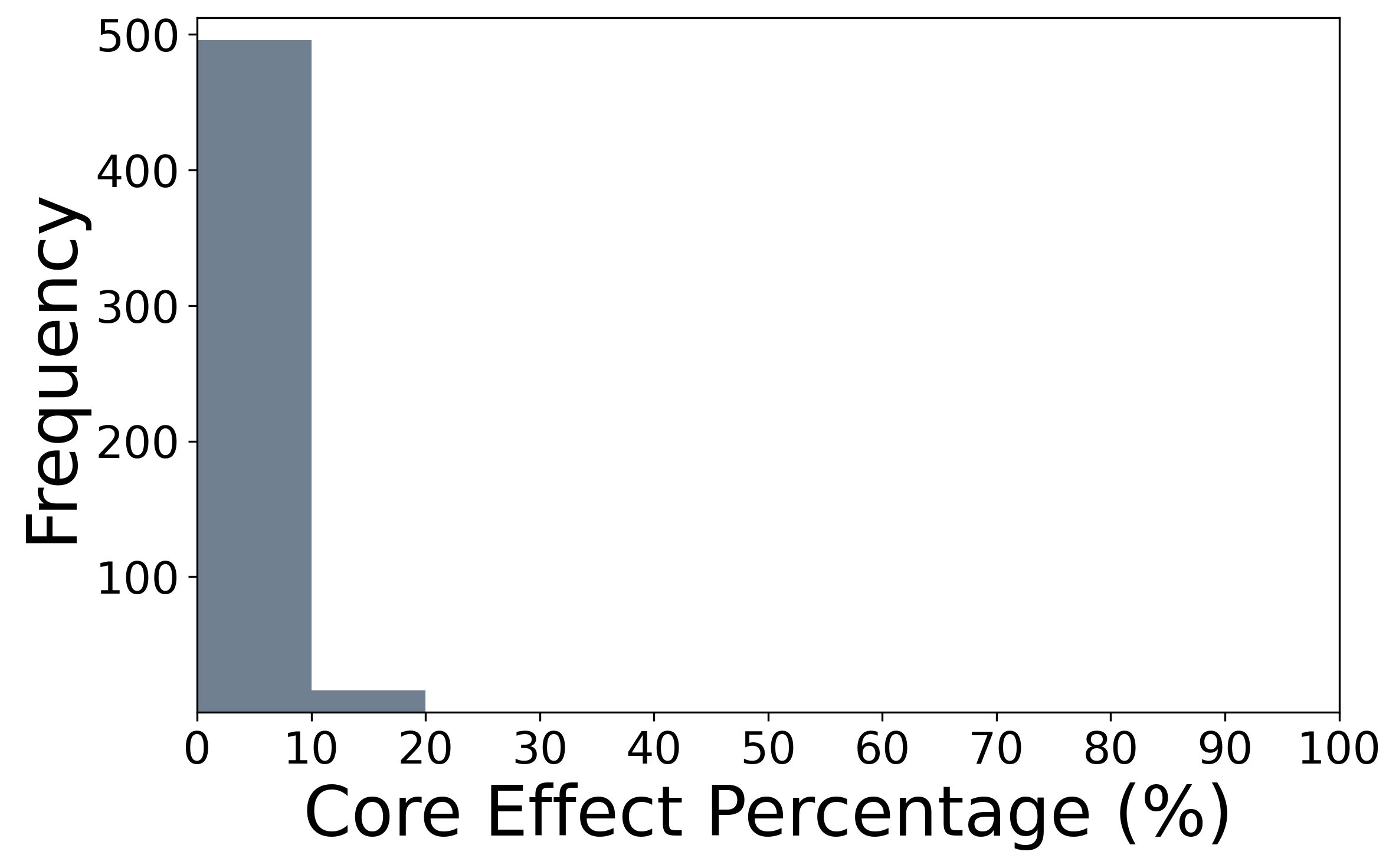}
  \caption{\(DAR_{Spu}\)}
  \label{fig:f_cep_dar_swap}
  \end{subfigure}
  \hspace*{\fill}   

\caption{Histogram of CEP values across 512 output features for various baseline methods. Refer to Section~\ref{section:feresults}, ~\ref{section:DAR_analysis}, and~\ref{section:experimental_setup} for a detailed analysis.}\label{fig:f_cep}
\end{figure}

We run experiments on the MNIST-CIFAR Dominoes~\cite{NEURIPS2020_6cfe0e61} dataset, which is constructed by concatenating MNIST~\cite{deng2012mnist} images with CIFAR-10~\cite{krizhevsky2009learning} images. The labels correspond to CIFAR-10 classes (core features) but are spuriously correlated with the MNIST digits (spurious features). This dataset is well-suited to our study because it supports controlled interventions on the input and provides straightforward interpretability via GradCAM~\cite{Selvaraju_2019} visualization. Details regarding the dataset and experimental setup are available in Appendix A and B, respectively. We assessed feature entanglement across various baseline methods using the CEP metric at different levels of the model’s representation: feature map (Figure~\ref{fig:map_cep}),  feature vector (Figure~\ref{fig:f_cep}), and output prediction (Table~\ref{Table:DAR_ablation}).

\begin{table*}[tb]
    \centering
    \caption{CEP values (\%) of the output prediction. Refer to Section \ref{section:feresults} and \ref{section:DAR_analysis} for a detailed analysis.}
    \resizebox{\textwidth}{!}{
    \begin{tabular}{c c c c c c c c}
        \toprule
        \textbf{Metric} & \(\mathbf{ERM_{Core}}\) & \(\mathbf{ERM}\) & \(\mathbf{DFR}\) & \(\mathbf{DFR_{CNN}}\) & \(\mathbf{DFR_{FC}}\) & \(\mathbf{DAR}\) & \(\mathbf{DAR_{Spu}}\)\\
        \toprule
        \textbf{CEP} & \(95.6 \pm 0.1\) & \(69.9 \pm 0.6\) & \(81.6 \pm 0.8\) & \(85.7 \pm 1.5\) & \(83.0 \pm 1.2\) & \(95.2 \pm 0.4\) & \(6.1 \pm 0.7\)\\
        \bottomrule
    \end{tabular}
    }
    \label{Table:DAR_ablation}
\end{table*}

\paragraph{\(ERM_{Core}\).} We train an ERM model on the dataset with a spurious strength of \(0\%\), ensuring the model learns only core features since the spurious features are not predictive. As expected, CEP values are close to \(100\%\) for all output levels: the feature maps (Figure \ref{fig:map_cep_erm_core}), features (Figure~\ref{fig:f_cep_erm_core}), and predictions (Table~\ref{Table:DAR_ablation}).

\paragraph{\(ERM\).} When trained on the dataset with the spurious correlation, the ERM model learns both core and spurious features since both are predictive. The feature maps exhibit significant entanglement (CEP $\approx 50\%$ in Figure~\ref{fig:map_cep_erm}). However, a more fine-grained pixel-level analysis reveals that different spatial locations within the same feature map often remain individually disentangled, with the pixel CEP values close to either \(0\%\) or \(100\%\) (Figure~\ref{fig:pixel_cep_erm}). The subsequent GAP layer discards this spatial information by averaging over the entire feature map, \textit{collapsing otherwise disentangled spatial activations into entangled output features} (Figure~\ref{fig:f_cep_erm}). Finally, output predictions depend on both the core and spurious features, as reflected in the low CEP score of $69.9\%$ (Table~\ref{Table:DAR_ablation}).

\paragraph{\(DFR\).} Because DFR operates on the already entangled ERM representation (Figure~\ref{fig:f_cep_erm}), it can only \emph{partially reduce reliance on spurious features rather than fully remove their influence} (CEP score of $81.6\%$ in Table~\ref{Table:DAR_ablation}). To further examine whether adapting only a single layer is sufficient for feature disentanglement~\cite{träuble2021disentangledrepresentationslearnedcorrelated}, we evaluate two variants: \(DFR_{CNN}\) and \(DFR_{FC}\). \(DFR_{CNN}\) additionally retrains the final CNN layer, leading to a slight improvement in CEP scores of the feature maps (Figure~\ref{fig:map_cep_cnn}). \(DFR_{FC}\) introduces an additional fully connected layer to transform the output features, which slightly improves the CEP scores of the output features (Figure~\ref{fig:f_cep_fc}). However, both variants remain insufficient to remove the influence of spurious features (CEP scores of $85.7\%$ and $83.0\%$ respectively in Table~\ref{Table:DAR_ablation}).

Beyond CEP, we provide an additional analysis in Appendix C and complementary visualizations in Appendix D, both of which corroborate our findings.
\section{Disentangling Features via Deep Attention Reweighting}
\label{section:DAR}
Building on our analysis in Section~\ref{section:entanglement}, we replace GAP with our proposed Deep Attention Reweighting (DAR), an attention-based aggregation mechanism that can adaptively assign importance weights to different spatial locations based on task relevance. This allows DAR to selectively emphasize core features and suppress spurious ones while they remain spatially separable within the feature maps. In this section, we formally introduce our approach, outline the design considerations, and present experimental evidence demonstrating its effectiveness.

\subsection{Motivation: Beyond GAP for Feature Aggregation}
\label{section:darmotivation}
In CNNs, the GAP layer is widely used for feature aggregation by uniformly averaging each feature map across all spatial locations. However, as shown in Section~\ref{section:entanglement}, this uniform weighting indiscriminately aggregates both core and spurious features distributed across spatial locations of feature maps, resulting in feature entanglement that undermines post-hoc reweighting strategies. To address this limitation, we seek an aggregation mechanism that fulfills two criteria:
\begin{enumerate}
    \item \textbf{Spatial weighting}: The aggregation mechanism should assign different \emph{importance weights to different spatial locations} within each feature map, allowing the model to emphasize task-relevant regions while suppressing irrelevant or spurious ones.
    \item \textbf{Adaptive weighting}: The importance weights should not be fixed learnable parameters, because the \emph{spatial locations of core and spurious features vary across inputs}. Instead, the weights should be input-conditional, yielding per-example importance maps specific to each input.
\end{enumerate}

\subsection{Attention-Based Feature Aggregation}
To satisfy the spatial and adaptive weighting criteria outlined above, we replace the GAP layer with an attention-based aggregation mechanism. Rather than uniformly averaging over all spatial positions, the model learns to assign input-dependent importance weights to different spatial locations of the feature map.

Let \(\mathbf{A}_i \in \mathbb{R}^{d \times H \times W}\) denote the activation map for input \(\mathbf{x}_i\), where $d$ is the number of channels and $H \times W$ is the spatial resolution. GAP aggregates each channel $j \in [d]$ across spatial locations to produce the resulting aggregated representation $\mathbf{h}_i \in \mathbb{R}^d$:
\begin{equation}
    \label{eqn:gap}
    \mathbf{h}_i[j] = \sum^H_{h=1} \sum^W_{w=1} \frac{1}{H*W} \mathbf{A}_i[j, h, w].
\end{equation}

We replace this with an attention-weighted aggregation:
\begin{equation}
    \label{eqn:dar}
    \mathbf{h}_i[j]\ = \sum^H_{h=1} \sum^W_{w=1} \mathbf{a}_i[j,h,w]  \mathbf{A}_i[j, h, w],
\end{equation}
where \(\mathbf{a}_i = f_{\mathrm{att}}(\mathbf{A}_i) \in \mathbb{R}^{d \times H \times W}\) denotes the attention weight assigned to each spatial location in each feature map while \(f_{\mathrm{att}}\) denotes the attention module. This attention-based aggregation mechanism satisfies the two aggregation criteria introduced in Section~\ref{section:darmotivation}: (1) spatial weighting via attention weights $\mathbf{a}_i[j, h, w]$, and (2) adaptive weighting since the $\mathbf{a}_i[j, h, w]$ is computed based on $\mathbf{A}_i$.

\subsection{Architectural Design of Attention Mechanism}
\label{section:design}
We adopt the scaled dot-product attention mechanism~\cite{NIPS2017_3f5ee243}, with the following design considerations tailored for post-hoc retraining on a small target set:

\begin{enumerate}
    \item \textbf{Query Vectors:} We use multiple learned query vectors, each of which independently computes attention weights over spatial locations of the feature map, allowing DAR to capture a diverse set of distinct core-feature patterns.
    \item \textbf{Positional Encoding:} We omit the use of positional encodings, which are typically added to inject information about the spatial position of each pixel. This preserves spatial invariance, ensuring that the attention weights depend only on the input content rather than their position.
    \item \textbf{Attention Depth:} We employ a two-layer configuration, where the learned query vectors are first updated through a contextual transformation to align more closely with input-specific feature distributions before being utilized in the attention computation, yielding a more adaptive weighting mechanism.
    \item \textbf{Multi-Headed Attention:} We utilize a multi-headed attention setup, where each attention head independently learns a set of attention weights to focus on distinct spatial regions. This allows for a more expressive aggregation mechanism that can capture features encoded across different feature maps and at different spatial locations.
\end{enumerate}
Refer to Section~\ref{sec:ablation} for the ablation results that empirically validate our proposed attention architecture and Appendix E for the full mathematical formulation of our attention architecture.
\subsection{Deep Attention Reweighting Algorithm}
The application of DAR involves three steps. First, we train an ERM model on the training set to obtain the backbone feature extractor. Since both core and spurious features are predictive, the resulting representation encodes both. Second, we replace GAP with the proposed attention-based aggregation module. This replaces uniform pooling with selective aggregation at the feature map level, where features are spatially disentangled. Finally, we fine-tune only the attention module and the classification head on a small, balanced target dataset where the spurious features are no longer predictive. This enables post-hoc mitigation of spurious correlations by selectively emphasizing core features and suppressing spurious ones during feature aggregation.

\subsection{Experimental Validation and Analysis}
\label{section:DAR_analysis}

To assess whether DAR effectively mitigates feature entanglement, we apply the same diagnostic tools and experimental setup in Section~\ref{section:entanglement}. As shown in Figure~\ref{fig:f_cep_dar}, DAR predominantly captures the core features while suppressing the spurious ones. Consequently, its predictions rely primarily on the core signal, achieving an output prediction CEP score of $\mathbf{95.2 \pm 0.4\%}$ (Table~\ref{Table:DAR_ablation}). To further examine the flexibility of the attention mechanism, we train \(DAR_{Spu}\), a variant that predicts the spurious attributes instead. In this configuration, the output feature vector relies predominantly on spurious features (Figure~\ref{fig:f_cep_dar_swap}), with an output prediction CEP score of $\mathbf{6.1 \pm 0.7\%}$ (Table~\ref{Table:DAR_ablation}). These results demonstrate the effectiveness of the attention-based feature-aggregation architecture in \emph{preserving core features while removing spurious ones} by selecting task-relevant spatial locations. This is supported by the additional entanglement metrics in Appendix C and GradCAM visualizations in Appendix D.
\section{Experiments}
\label{section:experiments}
We evaluate DAR across six spurious correlation benchmark datasets against competitive baselines and across diverse experimental configurations. The results show that DAR consistently outperforms the baselines, demonstrating its robustness in mitigating spurious correlations. We also present targeted ablation studies that characterize the source of DAR’s performance gains.

\subsection{Experimental Setup}
\label{section:experimental_setup}
\paragraph{Datasets.} We evaluate on six widely used spurious-correlation image benchmarks spanning diverse spurious structures. MNIST--CIFAR Dominoes~\cite{NEURIPS2020_6cfe0e61} exhibits clear \emph{spatial separation} between core and spurious regions; Waterbirds~\cite{welinder2010cub200, Sagawa2020Distributionally} captures \emph{background bias}, where background context contains the spurious features while the foreground bird is the core feature; Spuco-Animals~\cite{joshi2023mitigating} extends this setting to a more challenging \emph{4-class} classification task; Spawrious~\cite{lynch2023spawrious} contains photorealistic images \emph{artificially generated} by a text-to-image model; CelebA~\cite{https://doi.org/10.48550/arxiv.1411.7766, Sagawa2020Distributionally} has \emph{partial overlap} between core and spurious attributes; and UrbanCars~\cite{li2023whacamoledilemmashortcutscome} is a \emph{many-to-one} setting with multiple spurious cues predictive of the label. Additional details on the datasets are provided in Appendix A.

\paragraph{Baselines.} We compare DAR against a set of widely used competitive baseline methods for mitigating spurious correlations. Empirical Risk Minimization (ERM) trains the model directly on the full dataset without any additional techniques and serves as the naïve reference point. Just Train Twice (JTT)~\cite{DBLP:journals/corr/abs-2107-09044}, Correct-N-Contrast (CNC)~\cite{https://doi.org/10.48550/arxiv.2203.01517}, and Conditional Value at Risk Distributionally Robust Optimization (CVaR)~\cite{levy2020largescalemethodsdistributionallyrobust} do not require any additional information and infer the spurious attribute from the training loss, which is then used to regularize the training. The Reweight (RW), Subsample (SUB), and Group Distributionally Robust Optimization (gDRO)~\cite{Sagawa2020Distributionally} methods require access to the spurious attributes, using them to balance the training distribution to remove the statistical correlation between the spurious attributes and the labels. Deep Feature Reweighting (DFR)~\cite{https://doi.org/10.48550/arxiv.2204.02937} and our proposed DAR require a small, balanced target dataset for post-hoc reweighting. For a fair comparison, methods that do not use the target dataset will have it included in the training set.

\paragraph{Evaluation.} Consistent with the spurious-correlation literature~\cite{Sagawa2020Distributionally, yang2023change, ye2024spurious}, the minority-group accuracy is treated as the \emph{primary benchmark for robustness}. Subgroups are defined by the combination of the target label and a spurious attribute. The overrepresentation of a particular subgroup creates a statistical correlation between the corresponding spurious attribute and target label. Consequently, the majority group comprises subgroups in which the spurious correlation holds, whereas the minority group comprises subgroups in which it does not. Therefore, the majority-group accuracy, and by extension the average accuracy, is highly misleading as it may \emph{reflect successful learning of the spurious shortcut} rather than the core features. In contrast, the minority-group accuracy evaluates the model on \emph{datapoints where the spurious correlations do not hold}. We report the mean and standard deviation over three random seeds.

\paragraph{Training Setup.} The main experiment uses a randomly initialized ResNet18 model~\cite{he2015deep}. This random initialization isolates each method’s ability to learn feature representations independently of any pre-trained features from datasets such as ImageNet. The models are trained for \(200\) epochs to ensure convergence. We follow the hyperparameter-tuning setup described in \cite{yang2023change}, where the validation dataset is used to select the optimal hyperparameters for each method through grid search. Additional details of the experimental setup are provided in Appendix B.

\subsection{Main Results: Generalization Under Spurious Correlations}
\label{section:main_results}

\begin{table}[tb]
    \caption{Minority-Group Test Accuracy (\%) Across Datasets}
    \label{Table:main_results}
    \centering
    \resizebox{\textwidth}{!}{
    \begin{tabular}{c c c c c c c c c c}
        \toprule
        \textbf{Dataset} & \multicolumn{9}{c}{\textbf{Method}} \\
        \cmidrule(lr){2-10}
        & \textbf{ERM} & \textbf{JTT} & \textbf{CNC} & \textbf{CVaR} & \textbf{SUB} & \textbf{RW} & \textbf{gDRO} & \textbf{DFR} & \textbf{DAR} \\
        \midrule
        \textbf{Dominoes}      & $67.5\pm1.6$ & $69.4 \pm 0.5$ & $70.4 \pm 0.6$ & $68.9 \pm 1.1$ & $72.9 \pm 0.9$ & $76.1 \pm 1.6$ & $\underline{78.9 \pm 1.9}$ & $75.6 \pm 0.8$ & $\mathbf{79.0 \pm 0.4}$ \\
        \textbf{Spawrious}     & $89.2 \pm 1.0$ & $93.0 \pm 1.4$ &  $93.2\pm 1.3$ & $95.2 \pm 0.8$ & $\mathbf{98.8 \pm 0.5}$ & $95.8 \pm 1.1$ & $95.0 \pm 1.0$ & $94.6 \pm 1.4$ & $\underline{98.6 \pm 0.5}$ \\
        \textbf{Waterbirds}        & $67.4 \pm 0.7$ & $68.5 \pm 0.9$ & $69.2 \pm 0.3$ & $67.7 \pm 0.5$ & $\underline{75.1 \pm 0.6}$ & $75.0 \pm 0.9$ & $\mathbf{76.7 \pm 0.4}$ & $73.3 \pm 0.5$ & $74.6 \pm 0.7$ \\
        \textbf{Spuco}         & $39.8 \pm 1.7$ & $40.2 \pm 1.4$ & $42.1 \pm 1.1$ & $45.7 \pm 0.4$ & $59.2 \pm 0.6$ & $\underline{65.4 \pm 2.0}$ & $65.3 \pm 1.2$ & $63.2 \pm 1.0$ & $\mathbf{67.1 \pm 1.7}$ \\
        \textbf{CelebA}        & $90.8 \pm 2.1$ & $91.1 \pm 1.2$ & $92.3 \pm 1.1$ & $93.9 \pm 1.1$ & $94.7 \pm 0.5$ & $92.8 \pm 0.6$ & $95.0 \pm 0.9$ & $\underline{96.2 \pm 1.3}$ & $\mathbf{97.4 \pm 0.5}$  \\
        \textbf{UrbanCars}     & $73.0 \pm 1.0$ & $76.0 \pm 1.1$ & $75.4 \pm1.3$ & $73.9 \pm 0.4$ & $81.6 \pm 0.7$ & $79.4 \pm 0.6$ & $82.2 \pm 0.6$ & $\underline{87.1 \pm 0.5}$ & $\mathbf{89.7 \pm 1.0}$ \\
        \bottomrule
    \end{tabular}
    }
\end{table}

\begin{table}[tb]
    \caption{Minority-Group Test Accuracy (\%) Across Datasets and Configurations}
    \label{Table:additional_results}
    \centering
    \resizebox{\textwidth}{!}{
    \begin{tabular}{l l c c c c c c c c c}
        \toprule
        \multirow{2}{*}{\textbf{Dataset}} & \multirow{2}{*}{\textbf{Config}} &
        \multicolumn{9}{c}{\textbf{Method}} \\
        \cmidrule(lr){3-11}
        & & \textbf{ERM} & \textbf{JTT} & \textbf{CNC} & \textbf{CVaR} & \textbf{SUB} & \textbf{RW} & \textbf{gDRO} & \textbf{DFR} & \textbf{DAR} \\
        \midrule
        \multirow{3}{*}{\textbf{Dominoes}}
            & ResNet-50  & $67.4 \pm 1.7$ & $70.7 \pm 0.6$ & $70.5 \pm 1.1$ & $70.8 \pm 0.6$ & $72.0 \pm 1.3$ & $73.5 \pm 1.0$ & $73.3 \pm 1.2$ & $\underline{74.1 \pm 1.2}$ & $\mathbf{78.1 \pm 0.6}$ \\
            & 90\%       & $75.5 \pm 1.0$ & $76.8 \pm 0.4$ & $77.3 \pm 1.2$ & $77.9 \pm 0.8$ & $79.6 \pm 1.4$ & $79.1 \pm 0.7$ & $79.0 \pm 0.4$ & $\underline{81.7 \pm 1.8}$ & $\mathbf{82.6 \pm 1.7}$ \\
            & Pretrained & $81.9 \pm 0.3$ & $81.7 \pm 1.2$ & $83.7 \pm 2.3$ & $81.4 \pm 0.7$ & $84.3 \pm 0.5$ & $\underline{87.6 \pm 0.5}$ & $85.2 \pm 0.7$ & $87.6 \pm 0.8$ & $\mathbf{87.9 \pm 1.4}$ \\
        \addlinespace
        \multirow{3}{*}{\textbf{Spawrious}}
          & ResNet-50  & $88.8 \pm 0.5$ & $89.8 \pm 1.6$ & $93.4 \pm 0.8$ & $90.2 \pm 0.4$ & $\mathbf{96.6 \pm 1.6}$ & $95.8 \pm 0.7$ & $94.4 \pm 0.7$ & $95.0 \pm 0.6$ & $\underline{96.2 \pm 0.9}$ \\
          & 90\%       & $93.0 \pm 1.0$ & $95.0 \pm 1.5$ & $94.0 \pm 0.5$ & $96.0 \pm 1.7$ & $\mathbf{99.0 \pm 0.7}$ & $96.2 \pm 0.4$ & $96.4 \pm 0.6$ & $97.0 \pm 0.8$ & $\underline{98.8 \pm 0.6}$ \\
          & Pretrained & $98.2 \pm 0.8$ & $98.4 \pm 0.8$ & $98.0 \pm 0.4$ & $98.2 \pm 0.4 $ & $\underline{99.4 \pm 0.6}$ & $99.0 \pm 0.5$ & $98.7 \pm 0.9$ & $97.8 \pm 0.7$ & $\mathbf{99.4 \pm 0.4}$ \\
        \addlinespace
        \multirow{3}{*}{\textbf{Waterbirds}}
          & ResNet-50  & $63.6 \pm 0.3$ & $64.0 \pm 0.6$ & $67.1 \pm 1.3$ & $68.9 \pm 1.0$ & $74.2 \pm 1.8$ & $73.1 \pm 0.4$ & $73.5 \pm 1.1$ & $\underline{74.8 \pm 0.6}$ & $\mathbf{77.7 \pm 0.6}$ \\
          & 90\%       & $73.5 \pm 0.5$ & $74.7 \pm 0.8$ & $76.3 \pm 0.8$  & $75.7 \pm 0.5$& $\mathbf{85.4 \pm 0.4}$ & $85.2 \pm 1.6$ & $83.7 \pm 1.6$ & $84.3 \pm 0.7$ & $\underline{85.3 \pm 0.4}$ \\
          & Pretrained & $86.7 \pm 0.4$ & $88.1 \pm 0.4$ & $88.0 \pm 0.8$ & $86.5 \pm 1.4$ & $\mathbf{92.8 \pm 0.6}$ & $90.5 \pm 0.9$ & $91.7 \pm 0.7$ & $90.3 \pm 0.4$ & $\underline{92.0\pm 0.6}$ \\
        \addlinespace
        \multirow{3}{*}{\textbf{Spuco}}
          & ResNet-50  & $41.6 \pm 0.6$ & $44.8 \pm 0.4$ & $43.4 \pm 0.5$ & $47.8 \pm 1.0$ &$61.1 \pm 0.4$& $\underline{66.4 \pm 0.6}$ & $65.9 \pm 0.4$ & $64.3 \pm 0.7$ & $\mathbf{68.9 \pm 0.5}$ \\
          & 90\%       & $50.5 \pm 0.6$ & $57.7 \pm 0.8$ & $53.5 \pm 0.5$ & $53.2 \pm 0.7$ & $65.0 \pm 0.4$ & $\underline{67.3 \pm 0.9}$ & $66.9 \pm 0.6$ & $65.2 \pm 1.3$ & $\mathbf{70.9 \pm 0.4}$ \\
          & Pretrained & $59.9 \pm 0.3$ & $62.8 \pm 0.6$ & $61.7 \pm 0.8$ & $62.2 \pm 1.3$ & $76.5 \pm 0.8$ & $76.8 \pm 0.4$ & $78.3 \pm 0.7$ & $\underline{84.0 \pm 0.6}$ & $\mathbf{87.9 \pm 1.0}$ \\
        \addlinespace
        \multirow{3}{*}{\textbf{CelebA}}
          & ResNet-50  & $89.8 \pm 0.8$ &$ 90.7 \pm 0.3$ & $92.2 \pm 2.1$ & $92.6 \pm 0.4$ & $94.1 \pm 0.4$ & $92.3 \pm0.5$ & $93.2 \pm 1.5$  & $\underline{96.3 \pm 0.4}$ & $\mathbf{97.1 \pm 0.7}$  \\
          & 90\%       & $91.2 \pm 0.9$ & $92.3 \pm 0.9$ & $92.6 \pm 1.6$ & $95.0 \pm 0.9$ & $95.1 \pm 0.6$ & $94.6 \pm 1.5$ & $95.5 \pm 0.7$ & $\underline{97.5 \pm 0.6}$ & $\mathbf{98.2 \pm 0.5}$  \\
          & Pretrained & $91.6 \pm 1.3$ & $92.7 \pm 0.5$ & $94.2 \pm 0.8$ & $95.2 \pm 0.5$ & $96.3 \pm 0.3$ & $96.7 \pm 1.4$ & $96.4 \pm 0.5$ & $\underline{98.0 \pm 0.8}$ & $\mathbf{98.5 \pm 0.7}$  \\
        \addlinespace
        \multirow{3}{*}{\textbf{UrbanCars}}
            & ResNet-50  & $65.9 \pm 1.0$ & $67.2 \pm 1.2$ & $68.1 \pm 1.5$ & $66.4 \pm 1.2$ & $72.6 \pm 1.2$ & $77.0 \pm 1.4$ & $75.4 \pm 0.6$ & $\underline{79.9 \pm 0.8}$ & $\mathbf{84.1 \pm 1.3}$ \\
            & 90\%       & $83.3 \pm 0.4$ & $85.1\pm0.5$ & $84.8\pm1.3$ & $83.7\pm1.6$ & $86.6\pm1.3$ & $87.3\pm0.9$ & $\underline{89.7\pm0.5}$ & $88.2\pm0.8$ & $\mathbf{91.3 \pm 1.0}$ \\
            & Pretrained & $87.3 \pm 0.6$ & $88.5 \pm 1.4$ & $89.5 \pm 0.7$ & $87.8 \pm 1.1$ & $\mathbf{94.1 \pm 0.8}$ & $91.2 \pm 0.5$ & $92.3 \pm 0.5$ & $91.2 \pm 1.3$ & $\underline{93.7 \pm 0.5}$ \\
        \bottomrule
    \end{tabular}
    }
\end{table}

Table~\ref{Table:main_results} shows a clear performance hierarchy across method families. Firstly, ERM performs worst, illustrating how standard empirical risk minimization readily exploits spurious correlations in the training distribution and consequently fails to generalize to minority-group examples where those correlations break. Secondly, methods that infer spurious structure from training dynamics (JTT, CNC, and CVaR) generally improve over ERM, but their gains are modest, highlighting the limitations of approaches that do not rely on additional supervision signals. Thirdly, group-aware methods (SUB, RW, and gDRO), which explicitly use spurious-attribute annotations to rebalance or regularize training, achieve larger improvements, underscoring the value of leveraging spurious-group information to mitigate shortcut learning. Finally, despite its simplicity, DFR remains a strong post-hoc baseline: it is consistently competitive and, in several settings, even surpasses group-aware methods. This highlights the effectiveness of adapting using only a small target set, which requires minimal extra information and relaxes the requirement of spurious-attribute annotations.

Across all datasets in Table~\ref{Table:main_results}, \emph{DAR consistently outperforms DFR}, highlighting the robustness of the improvement gains across benchmarks. This improvement is consistent with the feature-entanglement analysis in Section~\ref{section:entanglement}: while DFR is a strong post-hoc adaptation baseline, its effectiveness is ultimately constrained by operating on an already collapsed feature vector in which core and spurious information remain entangled. By contrast, DAR replaces GAP with an attention-based aggregation at the feature-map level, where the core and spurious features are more spatially separable, enabling more selective emphasis of core features and suppression of spurious ones (Section~\ref{section:DAR}). Finally, this increased reliance on core features ultimately translates into more robust performance. Importantly, this observed improvement across all datasets suggests that DAR does not require that the core/spurious regions be fully disjoint. As long as \emph{core and spurious signals exhibit any spatial heterogeneity} (e.g., clear spatial separation, coarse localization differences, or partial spatial overlap), as is often the case in real-world datasets, DAR can \emph{exploit spatial structure to improve performance}. Moreover, DAR outperforms many baselines and is \emph{often superior to the strongest group-aware methods}, further supporting the effectiveness of post-hoc adaptation for mitigating spurious correlations.

Table~\ref{Table:additional_results} shows that the trends observed in the main benchmark setting remain consistent across diverse experimental configurations. 
These settings vary several of the most consequential aspects of the experimental setup and reflect common variations encountered in practice. Specifically, increasing the backbone from ResNet-18 to ResNet-50 tests whether DAR remains effective under a deeper, higher-capacity architecture; using pretrained initialization tests whether its gains persist when the model begins from a stronger feature prior; and reducing the spurious correlation strength from 95\% to 90\% evaluates performance under a less extreme shortcut regime. Across all datasets and realistic, high-impact variations, DAR maintains a consistent advantage over DFR and many other baselines, providing strong empirical evidence for the \emph{generalizability of our proposed DAR in mitigating spurious correlations}.

\subsection{Ablation Results}
\label{sec:ablation}
To better characterize DAR’s performance, we conduct a set of targeted ablation studies on the challenging and representative Spuco-Animals benchmark.

\begin{figure}[tb]
    \noindent\makebox[\linewidth]{%
        \hfill
        \begin{subfigure}[t]{0.4\linewidth}
            \centering
            \includegraphics[width=\linewidth]{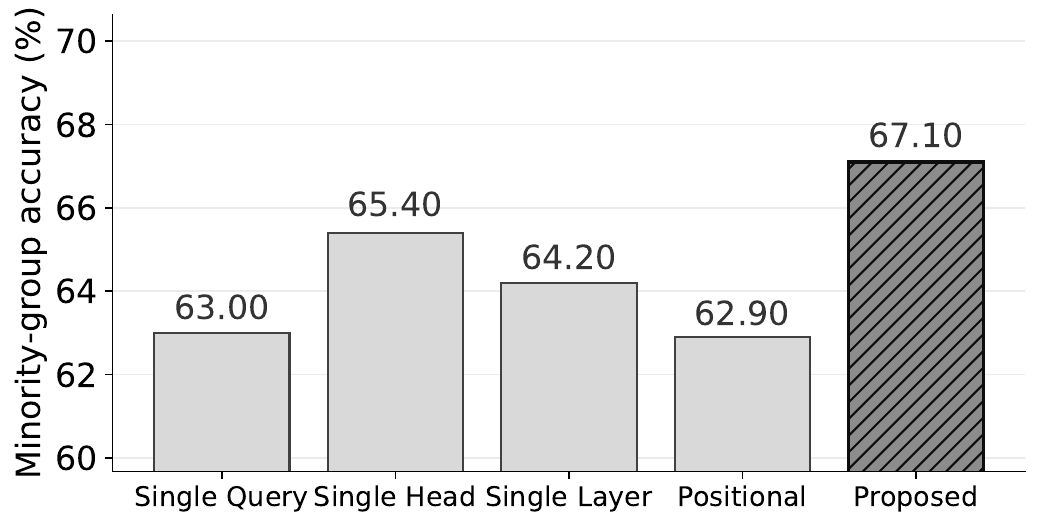}
            \caption{Attention Architecture.}
            \label{fig:post_hoc_arch_attention}
        \end{subfigure}%
        \hfill
        \begin{subfigure}[t]{0.35\linewidth}
            \centering
            \includegraphics[width=\linewidth]{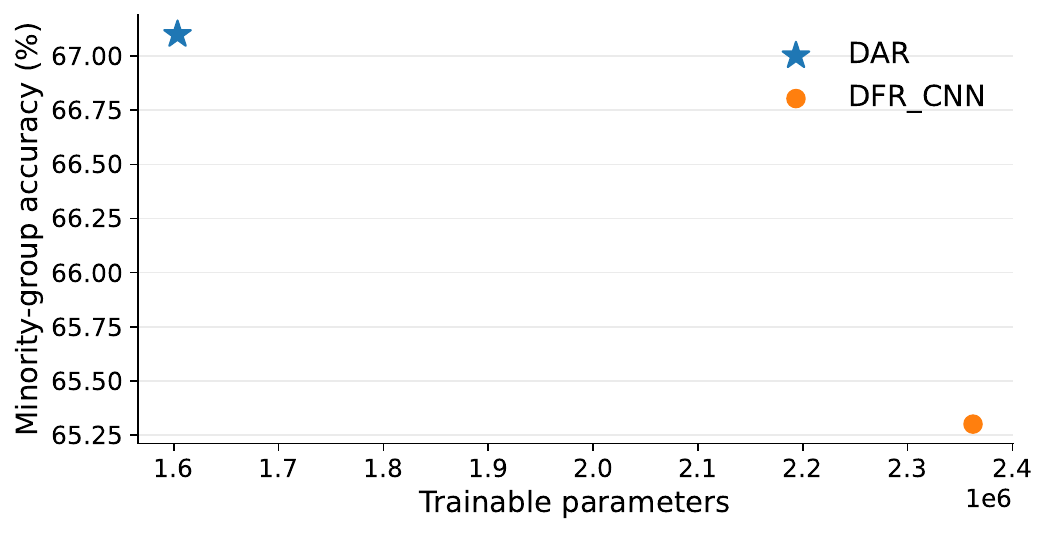}
            \caption{Post-Hoc Retraining Capacity.}
            \label{fig:post_hoc_arch_fc}
        \end{subfigure}%
        \hfill
    }
    \caption{\textbf{Post-hoc retraining architecture ablations.} (a) Attention architecture complexity ablation. (b) Post-Hoc retraining capacity ablation.}
    \label{fig:post_hoc_arch}
\end{figure}
\begin{figure}[tb]
    \centering
    \begin{subfigure}[t]{0.32\linewidth}
        \centering
        \includegraphics[width=\linewidth]{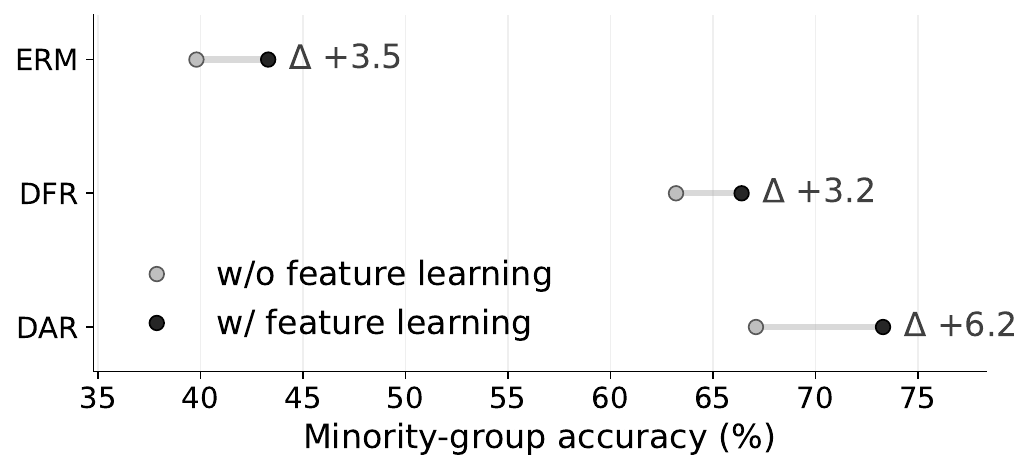}
        \caption{Feature Learning.}
        \label{fig:feature_learning}
    \end{subfigure}\hfill
    \begin{subfigure}[t]{0.32\linewidth}
        \centering
        \includegraphics[width=\linewidth]{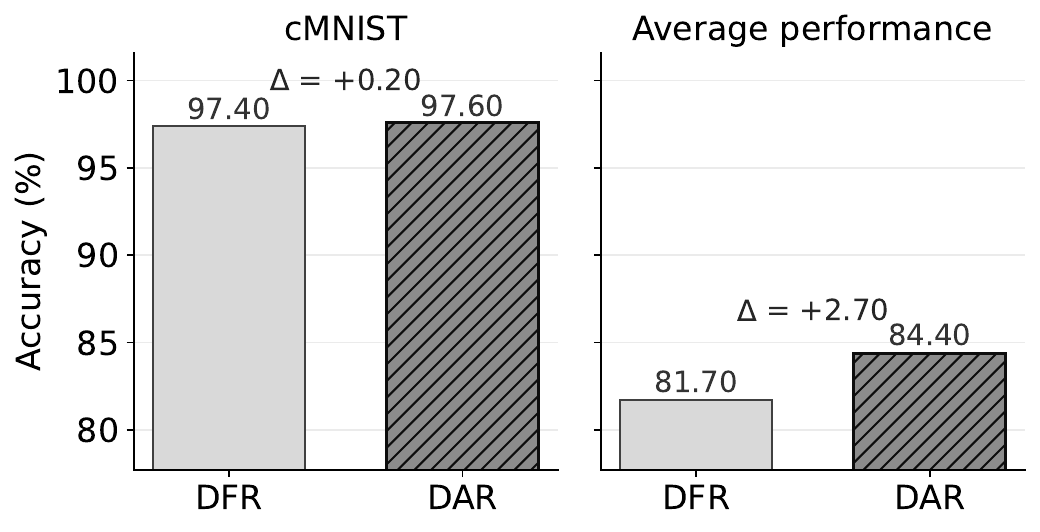}
        \caption{CMNIST.}
        \label{fig:CMNIST}
    \end{subfigure}
    \begin{subfigure}[t]{0.32\linewidth}
        \centering
        \includegraphics[width=\linewidth]{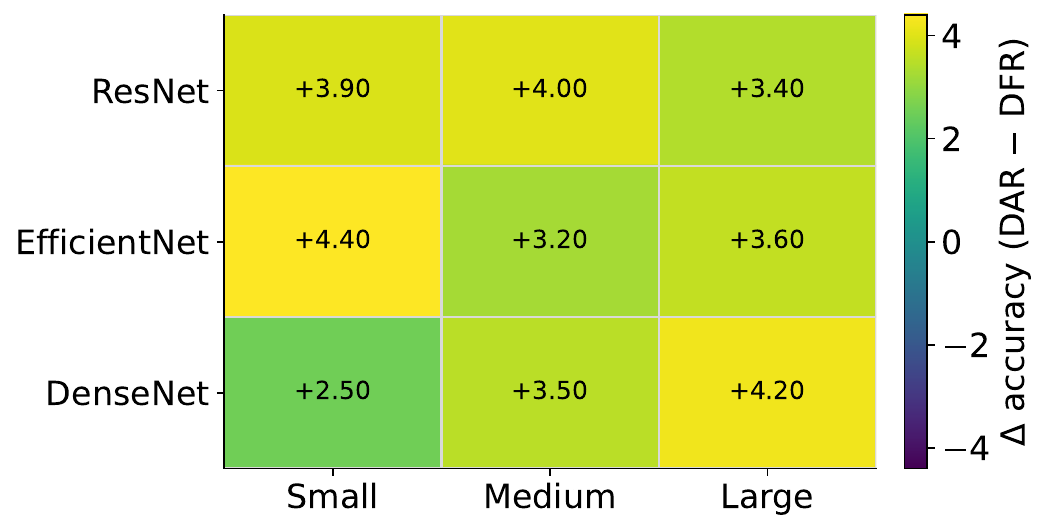}
        \caption{CNN Architecture.}
        \label{fig:architecture}
    \end{subfigure}
    \caption{Ablations for method characterization. (a) Feature learning compatibility. (b) Complete spatial overlap (CMNIST) robustness. (c) CNN architecture generality.}
    \label{fig:fig:char}
\end{figure}

\paragraph{Attention Architecture Ablation.}
Figure~\ref{fig:post_hoc_arch_attention} validates the attention-module design in Section~\ref{section:design} by ablating one component at a time from the proposed architecture. The proposed design performs best, and \emph{every simplification degrades performance}. These drops directly justify the use of multiple query vectors (to capture distinct core-feature patterns), multi-head attention (to model complementary attention patterns), a two-layer attention module (for more adaptive spatial weighting), and no positional encoding (to preserve spatial invariance).

\paragraph{Post-Hoc Retraining Capacity Ablation.}
Figure~\ref{fig:post_hoc_arch_fc} evaluates whether DAR’s gains can be explained solely by increased post-hoc model capacity by comparing it against $DFR_{CNN}$, a DFR variant that additionally retrains the final CNN layer, resulting in substantially more trainable parameters than DAR. Despite its greater adaptation capacity, $ DFR_{CNN}$ yields only modest improvements over DFR and still underperforms DAR. This shows that \emph{DAR’s advantage does not arise solely from post-hoc capacity}.

\paragraph{Compatibility with Feature Learning Methods.}
Figure~\ref{fig:feature_learning} evaluates the post-hoc DFR and DAR methods when applied to a model trained with the spectral decoupling loss~\cite{DBLP:journals/corr/abs-2011-09468}, a feature-learning technique designed to improve representations under spurious correlations. We observe that training the backbone with a feature-learning objective improves both DFR and DAR, with DAR remaining superior to DFR. These results show that \emph{post-hoc methods and feature-learning techniques are complementary}, as they optimize different aspects of the pipeline: feature learning improves the backbone representations, while post-hoc methods improve adaptation on top of the frozen features.

\paragraph{Performance under Complete Spatial Overlap.}
Figure~\ref{fig:CMNIST} evaluates DAR on Colored-MNIST (CMNIST)~\cite{arjovsky2020invariant}, a dataset in which the MNIST digits are colorized such that color is spuriously correlated with the digit labels. This yields an extreme setting in which the core and spurious features exhibit complete spatial overlap. In this setting, DAR achieves performance comparable to DFR. This result is consistent with DAR’s operating principle of exploiting any available spatial heterogeneity. When no such heterogeneity is present, DAR introduces no additional degradation and behaves similarly to DFR.

\paragraph{Performance across Base ERM Architectures.}
Figure~\ref{fig:architecture} evaluates the robustness of DAR across a diverse set of popular CNN backbones (ResNet~\cite{he2015deep}, EfficientNet~\cite{tan2020efficientnetrethinkingmodelscaling}, DenseNet~\cite{huang2018denselyconnectedconvolutionalnetworks}), spanning multiple capacity regimes (Small, Medium, Large). Across all evaluated architectures, DAR consistently outperforms DFR. These results highlight the \emph{broad generality of DAR across CNN architectures}.

\section{Limitations and Future Work}
\label{section:hit_limitations}

First, because DAR is a post-hoc method applied on top of a fixed backbone, its performance is constrained by the quality of the learned feature representation. Consequently, an insufficient representation may limit DAR's performance, even when it yields substantial gains. Indeed, DAR is naturally \emph{complementary to approaches for robust feature learning} \cite{DBLP:journals/corr/abs-2011-09468, huang2020selfchallengingimprovescrossdomaingeneralization, taghanaki2022masktune}. As demonstrated by the feature learning compatibility ablation in Section~\ref{sec:ablation}, integrating such techniques to improve representation quality can further enhance DAR's performance.

Second, DAR is limited in the extreme case of complete spatial overlap between the core and spurious signals. As shown in the spatial overlap ablation in Section~\ref{sec:ablation}, DAR provides limited additional benefit over DFR, although it introduces no further degradation. However, DAR does not require fully disjoint regions between the core and spurious: it provides the additional spatial dimensions for disentanglement, allowing it to exploit any spatial heterogeneity when present. To further improve DAR,  existing disentangled representation learning techniques \cite{9437964, Lee_2021_CVPR, pmlr-v162-kong22a, 10024368} can be incorporated into backbone training, which is complementary to  DAR.
\section{Conclusion}
\label{section:conclusion}
We introduce Deep Attention Reweighting, an attention-based feature aggregation mechanism that replaces Global Average Pooling in CNNs. DAR addresses feature entanglement, a key limitation of post-hoc Deep Feature Reweighting, by enabling selective aggregation across spatial regions within feature maps. Across various metrics, datasets, experimental setups, and ablations, we demonstrate DAR's ability to mitigate feature entanglement and spurious correlations. More broadly, this work highlights how architectural choices shape representation disentanglement and, in turn, influence its capacity to mitigate spurious correlations.

\bibliographystyle{splncs04}
\bibliography{main}

@String(CVPR  = {IEEE Conf. Comput. Vis. Pattern Recog.})

@String(ICCV  = {Int. Conf. Comput. Vis.})

@String(ECCV  = {Eur. Conf. Comput. Vis.})

@String(ICML  = {Int. Conf. Mach. Learn.})

@String(ICLR  = {Int. Conf. Learn. Represent.})

@inproceedings{https://doi.org/10.48550/arxiv.2110.14503,
  title={Simple Data Balancing Achieves Competitive Worst-Group-Accuracy},
  author={Idrissi, Badr and Arjovsky, Martin and Pezeshki, Mohammad and Lopez-Paz, David},
  booktitle={International Conference on Artificial Intelligence and Statistics (AISTATS)},
  year={2022},
  url={https://proceedings.mlr.press/v177/idrissi22a.html}
}

@article{Selvaraju_2019,
	doi = {10.1007/s11263-019-01228-7},
  
	url = {https://doi.org/10.1007/s11263-019-01228-7},
  
	year = 2019,
	month = {oct},
  
	publisher = {Springer Science and Business Media {LLC}},
  
	volume = {128},
  
	number = {2},
  
	pages = {336--359},
  
	author = {Ramprasaath R. Selvaraju and Michael Cogswell and Abhishek Das and Ramakrishna Vedantam and Devi Parikh and Dhruv Batra},
  
	title = {Grad-{CAM}: Visual Explanations from Deep Networks via Gradient-Based Localization},
  
	journal = {International Journal of Computer Vision}
}

@inproceedings{DBLP:journals/corr/abs-2107-09044,
  title={Just Train Twice: Improving Group Robustness without Training Group Information},
  author={Liu, Shuyang and Beery, Sara and Teney, Damien and Liu, Sijia and van den Hengel, Anton and Gould, Stephen},
  booktitle={International Conference on Machine Learning (ICML)},
  year={2021}
}

@inproceedings{https://doi.org/10.48550/arxiv.2203.01517,
  title={Correct-N-Contrast: A Contrastive Approach for Improving Robustness to Spurious Correlations},
  author={Zhang, Michael and Jia, Rui and Misra, Dipendra},
  booktitle={International Conference on Machine Learning (ICML)},
  year={2022},
  url={https://proceedings.mlr.press/v162/zhang22c.html}
}

@inproceedings{https://doi.org/10.48550/arxiv.1411.7766,
  title={Deep Learning Face Attributes in the Wild},
  author={Liu, Ziwei and Luo, Ping and Wang, Xiaogang and Tang, Xiaoou},
  booktitle={Proceedings of the IEEE International Conference on Computer Vision (ICCV)},
  year={2015},
  url={https://arxiv.org/abs/1411.7766}
}

@misc{lee2023diversify,
      title={Diversify and Disambiguate: Learning From Underspecified Data}, 
      author={Yoonho Lee and Huaxiu Yao and Chelsea Finn},
      year={2023},
      eprint={2202.03418},
      archivePrefix={arXiv},
      primaryClass={cs.LG}
}

@article{DBLP:journals/corr/abs-2011-09468,
  author       = {Mohammad Pezeshki and
                  S{\'{e}}kou{-}Oumar Kaba and
                  Yoshua Bengio and
                  Aaron C. Courville and
                  Doina Precup and
                  Guillaume Lajoie},
  title        = {Gradient Starvation: {A} Learning Proclivity in Neural Networks},
  journal      = {CoRR},
  volume       = {abs/2011.09468},
  year         = {2020},
  url          = {https://arxiv.org/abs/2011.09468},
  eprinttype    = {arXiv},
  eprint       = {2011.09468},
  bibsource    = {dblp computer science bibliography, https://dblp.org}
}

@misc{ye2024spurious,
      title={Spurious Correlations in Machine Learning: A Survey}, 
      author={Wenqian Ye and Guangtao Zheng and Xu Cao and Yunsheng Ma and Xia Hu and Aidong Zhang},
      year={2024},
      eprint={2402.12715},
      archivePrefix={arXiv},
      primaryClass={cs.LG}
}

@misc{nam2020learning,
      title={Learning from Failure: Training Debiased Classifier from Biased Classifier}, 
      author={Junhyun Nam and Hyuntak Cha and Sungsoo Ahn and Jaeho Lee and Jinwoo Shin},
      year={2020},
      eprint={2007.02561},
      archivePrefix={arXiv},
      primaryClass={cs.LG}
}

@misc{taghanaki2022masktune,
      title={MaskTune: Mitigating Spurious Correlations by Forcing to Explore}, 
      author={Saeid Asgari Taghanaki and Aliasghar Khani and Fereshte Khani and Ali Gholami and Linh Tran and Ali Mahdavi-Amiri and Ghassan Hamarneh},
      year={2022},
      eprint={2210.00055},
      archivePrefix={arXiv},
      primaryClass={cs.LG}
}

@InProceedings{pmlr-v119-ahuja20a,
  title = 	 {Invariant Risk Minimization Games},
  author =       {Ahuja, Kartik and Shanmugam, Karthikeyan and Varshney, Kush and Dhurandhar, Amit},
  booktitle = 	 {Proceedings of the 37th International Conference on Machine Learning},
  pages = 	 {145--155},
  year = 	 {2020},
  editor = 	 {III, Hal Daumé and Singh, Aarti},
  volume = 	 {119},
  series = 	 {Proceedings of Machine Learning Research},
  month = 	 {13--18 Jul},
  publisher =    {PMLR},
  url = 	 {https://proceedings.mlr.press/v119/ahuja20a.html}
}

@InProceedings{Lin_2022_CVPR,
    author    = {Lin, Yong and Dong, Hanze and Wang, Hao and Zhang, Tong},
    title     = {Bayesian Invariant Risk Minimization},
    booktitle = {Proceedings of the IEEE/CVF Conference on Computer Vision and Pattern Recognition (CVPR)},
    month     = {June},
    year      = {2022},
    pages     = {16021-16030}
}

@InProceedings{pmlr-v162-zhou22e,
  title = 	 {Sparse Invariant Risk Minimization},
  author =       {Zhou, Xiao and Lin, Yong and Zhang, Weizhong and Zhang, Tong},
  booktitle = 	 {Proceedings of the 39th International Conference on Machine Learning},
  pages = 	 {27222--27244},
  year = 	 {2022},
  editor = 	 {Chaudhuri, Kamalika and Jegelka, Stefanie and Song, Le and Szepesvari, Csaba and Niu, Gang and Sabato, Sivan},
  volume = 	 {162},
  series = 	 {Proceedings of Machine Learning Research},
  month = 	 {17--23 Jul},
  publisher =    {PMLR},
  url = 	 {https://proceedings.mlr.press/v162/zhou22e.html}
}

@misc{pagliardini2022agree,
      title={Agree to Disagree: Diversity through Disagreement for Better Transferability}, 
      author={Matteo Pagliardini and Martin Jaggi and François Fleuret and Sai Praneeth Karimireddy},
      year={2022},
      eprint={2202.04414},
      archivePrefix={arXiv},
      primaryClass={cs.LG}
}

@inproceedings{yang2023change,
author = {Yang, Yuzhe and Zhang, Haoran and Katabi, Dina and Ghassemi, Marzyeh},
title = {Change is hard: a closer look at subpopulation shift},
year = {2023},
publisher = {JMLR.org},
booktitle = {Proceedings of the 40th International Conference on Machine Learning},
articleno = {1652},
numpages = {39},
location = {Honolulu, Hawaii, USA},
series = {ICML'23}
}

@misc{joshi2023mitigating,
      title={Towards Mitigating Spurious Correlations in the Wild: A Benchmark and a more Realistic Dataset}, 
      author={Siddharth Joshi and Yu Yang and Yihao Xue and Wenhan Yang and Baharan Mirzasoleiman},
      year={2023},
      eprint={2306.11957},
      archivePrefix={arXiv},
      primaryClass={cs.LG}
}

@misc{lynch2023spawrious,
      title={Spawrious: A Benchmark for Fine Control of Spurious Correlation Biases}, 
      author={Aengus Lynch and Gbètondji J-S Dovonon and Jean Kaddour and Ricardo Silva},
      year={2023},
      eprint={2303.05470},
      archivePrefix={arXiv},
      primaryClass={cs.CV}
}

@misc{lopezpaz2016dependence,
      title={From Dependence to Causation}, 
      author={David Lopez-Paz},
      year={2016},
      eprint={1607.03300},
      archivePrefix={arXiv},
      primaryClass={stat.ML}
}

@InProceedings{pmlr-v202-liang23d,
  title = 	 {Accuracy on the Curve: On the Nonlinear Correlation of {ML} Performance Between Data Subpopulations},
  author =       {Liang, Weixin and Mao, Yining and Kwon, Yongchan and Yang, Xinyu and Zou, James},
  booktitle = 	 {Proceedings of the 40th International Conference on Machine Learning},
  pages = 	 {20706--20724},
  year = 	 {2023},
  editor = 	 {Krause, Andreas and Brunskill, Emma and Cho, Kyunghyun and Engelhardt, Barbara and Sabato, Sivan and Scarlett, Jonathan},
  volume = 	 {202},
  series = 	 {Proceedings of Machine Learning Research},
  month = 	 {23--29 Jul},
  publisher =    {PMLR},
  url = 	 {https://proceedings.mlr.press/v202/liang23d.html}
}

@misc{chen2023confidencebased,
      title={Confidence-Based Model Selection: When to Take Shortcuts for Subpopulation Shifts}, 
      author={Annie S. Chen and Yoonho Lee and Amrith Setlur and Sergey Levine and Chelsea Finn},
      year={2023},
      eprint={2306.11120},
      archivePrefix={arXiv},
      primaryClass={cs.LG}
}

@misc{lake2016building,
      title={Building Machines That Learn and Think Like People}, 
      author={Brenden M. Lake and Tomer D. Ullman and Joshua B. Tenenbaum and Samuel J. Gershman},
      year={2016},
      eprint={1604.00289},
      archivePrefix={arXiv},
      primaryClass={cs.AI}
}

@misc{marcus2018deep,
      title={Deep Learning: A Critical Appraisal}, 
      author={Gary Marcus},
      year={2018},
      eprint={1801.00631},
      archivePrefix={arXiv},
      primaryClass={cs.AI}
}

@misc{peters2015causal,
      title={Causal inference using invariant prediction: identification and confidence intervals}, 
      author={Jonas Peters and Peter Bühlmann and Nicolai Meinshausen},
      year={2015},
      eprint={1501.01332},
      archivePrefix={arXiv},
      primaryClass={stat.ME}
}

@misc{heinzedeml2018invariant,
      title={Invariant Causal Prediction for Nonlinear Models}, 
      author={Christina Heinze-Deml and Jonas Peters and Nicolai Meinshausen},
      year={2018},
      eprint={1706.08576},
      archivePrefix={arXiv},
      primaryClass={stat.ME}
}

@article{deng2012mnist,
  title={The mnist database of handwritten digit images for machine learning research},
  author={Deng, Li},
  journal={IEEE Signal Processing Magazine},
  volume={29},
  number={6},
  pages={141--142},
  year={2012},
  publisher={IEEE}
}

@Techreport{krizhevsky2009learning,
 author = {Krizhevsky, Alex and Hinton, Geoffrey},
 address = {Toronto, Ontario},
 institution = {University of Toronto},
 number = {0},
 publisher = {Technical report, University of Toronto},
 title = {Learning multiple layers of features from tiny images},
 year = {2009},
 title_with_no_special_chars = {Learning multiple layers of features from tiny images},
 url = {https://www.cs.toronto.edu/~kriz/learning-features-2009-TR.pdf}
}

@inproceedings{he2015deep,
  title={Deep Residual Learning for Image Recognition},
  author={He, Kaiming and Zhang, Xiangyu and Ren, Shaoqing and Sun, Jian},
  booktitle={Proceedings of the IEEE Conference on Computer Vision and Pattern Recognition (CVPR)},
  year={2016},
  pages={770--778},
  url={https://openaccess.thecvf.com/content_cvpr_2016/html/He_Deep_Residual_Learning_CVPR_2016_paper.html}
}

@InProceedings{KingBa15,
  author    = {Kingma, Diederik and Ba, Jimmy},
  booktitle = {International Conference on Learning Representations (ICLR)},
  title     = {Adam: A Method for Stochastic Optimization},
  year      = {2015},
  address   = {San Diego, CA, USA},
  optmonth  = {12},
}

@inproceedings{NEURIPS2023_265bee74,
 author = {LaBonte, Tyler and Muthukumar, Vidya and Kumar, Abhishek},
 booktitle = {Advances in Neural Information Processing Systems},
 editor = {A. Oh and T. Neumann and A. Globerson and K. Saenko and M. Hardt and S. Levine},
 pages = {11552--11579},
 publisher = {Curran Associates, Inc.},
 title = {Towards Last-layer Retraining for Group Robustness with Fewer Annotations},
 url = {https://proceedings.neurips.cc/paper_files/paper/2023/file/265bee74aee86df77e8e36d25e786ab5-Paper-Conference.pdf},
 volume = {36},
 year = {2023}
}

@InProceedings{pmlr-v202-qiu23c,
  title = 	 {Simple and Fast Group Robustness by Automatic Feature Reweighting},
  author =       {Qiu, Shikai and Potapczynski, Andres and Izmailov, Pavel and Wilson, Andrew Gordon},
  booktitle = 	 {Proceedings of the 40th International Conference on Machine Learning},
  pages = 	 {28448--28467},
  year = 	 {2023},
  editor = 	 {Krause, Andreas and Brunskill, Emma and Cho, Kyunghyun and Engelhardt, Barbara and Sabato, Sivan and Scarlett, Jonathan},
  volume = 	 {202},
  series = 	 {Proceedings of Machine Learning Research},
  month = 	 {23--29 Jul},
  publisher =    {PMLR},
  url = 	 {https://proceedings.mlr.press/v202/qiu23c.html}
}

@misc{locatello2020disentanglingfactorsvariationusing,
      title={Disentangling Factors of Variation Using Few Labels}, 
      author={Francesco Locatello and Michael Tschannen and Stefan Bauer and Gunnar Rätsch and Bernhard Schölkopf and Olivier Bachem},
      year={2020},
      eprint={1905.01258},
      archivePrefix={arXiv},
      primaryClass={cs.LG},
      url={https://arxiv.org/abs/1905.01258}, 
}

@article{bengio2014representationlearningreviewnew,
  title={Representation Learning: A Review and New Perspectives},
  author={Bengio, Yoshua and Courville, Aaron and Vincent, Pascal},
  journal={IEEE Transactions on Pattern Analysis and Machine Intelligence},
  year={2013},
  doi={10.1109/TPAMI.2013.50}
}

@inproceedings{kingma2022autoencodingvariationalbayes,
  title={Auto-Encoding Variational Bayes},
  author={Kingma, Diederik P. and Welling, Max},
  booktitle={International Conference on Learning Representations (ICLR)},
  year={2014},
  eprint={1312.6114},
  archivePrefix={arXiv},
  primaryClass={stat.ML},
  url={https://arxiv.org/abs/1312.6114}
}

@misc{träuble2021disentangledrepresentationslearnedcorrelated,
      title={On Disentangled Representations Learned From Correlated Data}, 
      author={Frederik Träuble and Elliot Creager and Niki Kilbertus and Francesco Locatello and Andrea Dittadi and Anirudh Goyal and Bernhard Schölkopf and Stefan Bauer},
      year={2021},
      eprint={2006.07886},
      archivePrefix={arXiv},
      primaryClass={cs.LG},
      url={https://arxiv.org/abs/2006.07886}, 
}

@misc{wu2020visualtransformerstokenbasedimage,
      title={Visual Transformers: Token-based Image Representation and Processing for Computer Vision}, 
      author={Bichen Wu and Chenfeng Xu and Xiaoliang Dai and Alvin Wan and Peizhao Zhang and Zhicheng Yan and Masayoshi Tomizuka and Joseph Gonzalez and Kurt Keutzer and Peter Vajda},
      year={2020},
      eprint={2006.03677},
      archivePrefix={arXiv},
      primaryClass={cs.CV},
      url={https://arxiv.org/abs/2006.03677}, 
}

@inproceedings{dosovitskiy2021imageworth16x16words,
  title={An Image is Worth 16x16 Words: Transformers for Image Recognition at Scale},
  author={Dosovitskiy, Alexey and Beyer, Lucas and Kolesnikov, Alexander and Weissenborn, Dirk and Zhai, Xiaohua and Unterthiner, Thomas and Dehghani, Mostafa and Minderer, Matthias and Heigold, Georg and Gelly, Sylvain and Uszkoreit, Jakob and Houlsby, Neil},
  booktitle={International Conference on Learning Representations (ICLR)},
  year={2021},
  url={https://openreview.net/forum?id=YicbFdNTTy}
}

@inproceedings{NIPS2017_3f5ee243,
 author = {Vaswani, Ashish and Shazeer, Noam and Parmar, Niki and Uszkoreit, Jakob and Jones, Llion and Gomez, Aidan N and Kaiser, \L ukasz and Polosukhin, Illia},
 booktitle = {Advances in Neural Information Processing Systems},
 editor = {I. Guyon and U. Von Luxburg and S. Bengio and H. Wallach and R. Fergus and S. Vishwanathan and R. Garnett},
 pages = {},
 publisher = {Curran Associates, Inc.},
 title = {Attention is All you Need},
 url = {https://proceedings.neurips.cc/paper_files/paper/2017/file/3f5ee243547dee91fbd053c1c4a845aa-Paper.pdf},
 volume = {30},
 year = {2017}
}

@ARTICLE{9716741,
  author={Han, Kai and Wang, Yunhe and Chen, Hanting and Chen, Xinghao and Guo, Jianyuan and Liu, Zhenhua and Tang, Yehui and Xiao, An and Xu, Chunjing and Xu, Yixing and Yang, Zhaohui and Zhang, Yiman and Tao, Dacheng},
  journal={IEEE Transactions on Pattern Analysis and Machine Intelligence}, 
  title={A Survey on Vision Transformer}, 
  year={2023},
  volume={45},
  number={1},
  pages={87-110},
  doi={10.1109/TPAMI.2022.3152247}}

@misc{pearl2012docalculusrevisited,
      title={The Do-Calculus Revisited}, 
      author={Judea Pearl},
      year={2012},
      eprint={1210.4852},
      archivePrefix={arXiv},
      primaryClass={cs.AI},
      url={https://arxiv.org/abs/1210.4852}, 
}

@ARTICLE{10579040,
  author={Wang, Xin and Chen, Hong and Tang, Si'ao and Wu, Zihao and Zhu, Wenwu},
  journal={IEEE Transactions on Pattern Analysis and Machine Intelligence}, 
  title={Disentangled Representation Learning}, 
  year={2024},
  volume={46},
  number={12},
  pages={9677-9696},
  doi={10.1109/TPAMI.2024.3420937}}

@misc{kumar2018variationalinferencedisentangledlatent,
      title={Variational Inference of Disentangled Latent Concepts from Unlabeled Observations}, 
      author={Abhishek Kumar and Prasanna Sattigeri and Avinash Balakrishnan},
      year={2018},
      eprint={1711.00848},
      archivePrefix={arXiv},
      primaryClass={cs.LG},
      url={https://arxiv.org/abs/1711.00848}, 
}

@misc{kim2019disentanglingfactorising,
      title={Disentangling by Factorising}, 
      author={Hyunjik Kim and Andriy Mnih},
      year={2019},
      eprint={1802.05983},
      archivePrefix={arXiv},
      primaryClass={stat.ML},
      url={https://arxiv.org/abs/1802.05983}, 
}

@inproceedings{higgins2017beta,
  title={Beta-VAE: Learning Basic Visual Concepts with a Constrained Variational Framework},
  author={Higgins, Irina and Matthey, Lo{\"i}c and Pal, Arka and Burgess, Christopher P and Glorot, Xavier and Botvinick, Matthew and Mohamed, Shakir and Lerchner, Alexander},
  booktitle={International Conference on Learning Representations (ICLR)},
  year={2017}
}

@ARTICLE{9947342,
  author={Carbonneau, Marc-André and Zaïdi, Julian and Boilard, Jonathan and Gagnon, Ghyslain},
  journal={IEEE Transactions on Neural Networks and Learning Systems}, 
  title={Measuring Disentanglement: A Review of Metrics}, 
  year={2024},
  volume={35},
  number={7},
  pages={8747-8761},
  doi={10.1109/TNNLS.2022.3218982}}

@misc{chen2019isolatingsourcesdisentanglementvariational,
      title={Isolating Sources of Disentanglement in Variational Autoencoders}, 
      author={Ricky T. Q. Chen and Xuechen Li and Roger Grosse and David Duvenaud},
      year={2019},
      eprint={1802.04942},
      archivePrefix={arXiv},
      primaryClass={cs.LG},
      url={https://arxiv.org/abs/1802.04942}, 
}

@misc{dupont2018learningdisentangledjointcontinuous,
      title={Learning Disentangled Joint Continuous and Discrete Representations}, 
      author={Emilien Dupont},
      year={2018},
      eprint={1804.00104},
      archivePrefix={arXiv},
      primaryClass={stat.ML},
      url={https://arxiv.org/abs/1804.00104}, 
}

@misc{kim2019relevancefactorvaelearning,
      title={Relevance Factor VAE: Learning and Identifying Disentangled Factors}, 
      author={Minyoung Kim and Yuting Wang and Pritish Sahu and Vladimir Pavlovic},
      year={2019},
      eprint={1902.01568},
      archivePrefix={arXiv},
      primaryClass={cs.LG},
      url={https://arxiv.org/abs/1902.01568}, 
}

@misc{burgess2018understandingdisentanglingbetavae,
      title={Understanding disentangling in $\beta$-VAE}, 
      author={Christopher P. Burgess and Irina Higgins and Arka Pal and Loic Matthey and Nick Watters and Guillaume Desjardins and Alexander Lerchner},
      year={2018},
      eprint={1804.03599},
      archivePrefix={arXiv},
      primaryClass={stat.ML},
      url={https://arxiv.org/abs/1804.03599}, 
}

@misc{higgins2018scanlearninghierarchicalcompositional,
      title={SCAN: Learning Hierarchical Compositional Visual Concepts}, 
      author={Irina Higgins and Nicolas Sonnerat and Loic Matthey and Arka Pal and Christopher P Burgess and Matko Bosnjak and Murray Shanahan and Matthew Botvinick and Demis Hassabis and Alexander Lerchner},
      year={2018},
      eprint={1707.03389},
      archivePrefix={arXiv},
      primaryClass={stat.ML},
      url={https://arxiv.org/abs/1707.03389}, 
}

@misc{mathieu2019disentanglingdisentanglementvariationalautoencoders,
      title={Disentangling Disentanglement in Variational Autoencoders}, 
      author={Emile Mathieu and Tom Rainforth and N. Siddharth and Yee Whye Teh},
      year={2019},
      eprint={1812.02833},
      archivePrefix={arXiv},
      primaryClass={stat.ML},
      url={https://arxiv.org/abs/1812.02833}, 
}

@inproceedings{wang2018nonlocalneuralnetworks,
  title={Non-local Neural Networks},
  author={Wang, Xiaolong and Girshick, Ross and Gupta, Abhinav and He, Kaiming},
  booktitle={Proceedings of the IEEE Conference on Computer Vision and Pattern Recognition (CVPR)},
  year={2018},
  pages={7794--7803},
  url={https://openaccess.thecvf.com/content_cvpr_2018/html/Wang_Non-Local_Neural_Networks_CVPR_2018_paper.html}
}

@inproceedings{touvron2021trainingdataefficientimagetransformers,
  title={Training data-efficient image transformers \& distillation through attention},
  author={Touvron, Hugo and Cord, Matthieu and Douze, Matthijs and Massa, Francisco and Sablayrolles, Alexandre and J{'e}gou, Herv{'e}},
  booktitle={International Conference on Machine Learning (ICML)},
  year={2021},
  url={https://proceedings.mlr.press/v139/touvron21a.html}
}

@inproceedings{liu2021swintransformerhierarchicalvision,
  title={Swin Transformer: Hierarchical Vision Transformer using Shifted Windows},
  author={Liu, Ze and Lin, Yutong and Cao, Yue and Hu, Han and Wei, Yixuan and Zhang, Zheng and Lin, Stephen and Guo, Baining},
  booktitle={Proceedings of the IEEE/CVF International Conference on Computer Vision (ICCV)},
  year={2021},
  url={https://openaccess.thecvf.com/content/ICCV2021/html/Liu_Swin_Transformer_Hierarchical_Vision_Transformer_using_Shifted_Windows_ICCV_2021_paper.html}
}

@misc{wang2021pyramidvisiontransformerversatile,
      title={Pyramid Vision Transformer: A Versatile Backbone for Dense Prediction without Convolutions}, 
      author={Wenhai Wang and Enze Xie and Xiang Li and Deng-Ping Fan and Kaitao Song and Ding Liang and Tong Lu and Ping Luo and Ling Shao},
      year={2021},
      eprint={2102.12122},
      archivePrefix={arXiv},
      primaryClass={cs.CV},
      url={https://arxiv.org/abs/2102.12122}, 
}

@misc{yuan2021tokenstotokenvittrainingvision,
      title={Tokens-to-Token ViT: Training Vision Transformers from Scratch on ImageNet}, 
      author={Li Yuan and Yunpeng Chen and Tao Wang and Weihao Yu and Yujun Shi and Zihang Jiang and Francis EH Tay and Jiashi Feng and Shuicheng Yan},
      year={2021},
      eprint={2101.11986},
      archivePrefix={arXiv},
      primaryClass={cs.CV},
      url={https://arxiv.org/abs/2101.11986}, 
}

@inproceedings{hu2019squeezeandexcitationnetworks,
  title={Squeeze-and-Excitation Networks},
  author={Hu, Jie and Shen, Li and Sun, Gang},
  booktitle={Proceedings of the IEEE Conference on Computer Vision and Pattern Recognition (CVPR)},
  year={2018},
  url={https://openaccess.thecvf.com/content_cvpr_2018/html/Hu_Squeeze-and-Excitation_Networks_CVPR_2018_paper.html}
}

@inproceedings{bello2020attentionaugmentedconvolutionalnetworks,
  title={Attention Augmented Convolutional Networks},
  author={Bello, Irwan and Zoph, Barret and Vaswani, Ashish and Shlens, Jonathon and Le, Quoc V.},
  booktitle={Proceedings of the IEEE/CVF International Conference on Computer Vision (ICCV)},
  year={2019},
  url={https://openaccess.thecvf.com/content_ICCV_2019/html/Bello_Attention_Augmented_Convolutional_Networks_ICCV_2019_paper.html}
}

@inproceedings{woo2018cbamconvolutionalblockattention,
  title={CBAM: Convolutional Block Attention Module},
  author={Woo, Sanghyun and Park, Jongchan and Lee, Joon-Young and Kweon, In So},
  booktitle={Computer Vision -- ECCV 2018},
  year={2018},
  pages={3--19},
  doi={10.1007/978-3-030-01234-2_1},
  url={https://openaccess.thecvf.com/content_ECCV_2018/html/Sanghyun_Woo_Convolutional_Block_Attention_Module_ECCV_2018_paper.html}
}

@inproceedings{jetley2018learnpayattention,
  title={Learn To Pay Attention},
  author={Jetley, Saumya and Lord, Nicholas A. and Lee, Namhoon and Torr, Philip H. S.},
  booktitle={International Conference on Learning Representations (ICLR)},
  year={2018},
  url={https://openreview.net/forum?id=HkG3SJZ1D}
}

@misc{ghosal2022visiontransformersrobustspurious,
      title={Are Vision Transformers Robust to Spurious Correlations?}, 
      author={Soumya Suvra Ghosal and Yifei Ming and Yixuan Li},
      year={2022},
      eprint={2203.09125},
      archivePrefix={arXiv},
      primaryClass={cs.CV},
      url={https://arxiv.org/abs/2203.09125}, 
}

@INPROCEEDINGS{10604648,
  author={Yue, Dengfeng and Zou, Jia and Jin, Xiaoxiao and Leng, Tuo},
  booktitle={2024 7th International Conference on Artificial Intelligence and Big Data (ICAIBD)}, 
  title={Causal Inference for Confounder-Purify Vision Transformers}, 
  year={2024},
  volume={},
  number={},
  pages={530-537},
  doi={10.1109/ICAIBD62003.2024.10604648}}

@misc{yang2021causalattentionvisionlanguagetasks,
      title={Causal Attention for Vision-Language Tasks}, 
      author={Xu Yang and Hanwang Zhang and Guojun Qi and Jianfei Cai},
      year={2021},
      eprint={2103.03493},
      archivePrefix={arXiv},
      primaryClass={cs.CV},
      url={https://arxiv.org/abs/2103.03493}, 
}

@misc{wang2021causalattentionunbiasedvisual,
      title={Causal Attention for Unbiased Visual Recognition}, 
      author={Tan Wang and Chang Zhou and Qianru Sun and Hanwang Zhang},
      year={2021},
      eprint={2108.08782},
      archivePrefix={arXiv},
      primaryClass={cs.CV},
      url={https://arxiv.org/abs/2108.08782}, 
}

@misc{huang2020selfchallengingimprovescrossdomaingeneralization,
      title={Self-Challenging Improves Cross-Domain Generalization}, 
      author={Zeyi Huang and Haohan Wang and Eric P. Xing and Dong Huang},
      year={2020},
      eprint={2007.02454},
      archivePrefix={arXiv},
      primaryClass={cs.CV},
      url={https://arxiv.org/abs/2007.02454}, 
}

@inproceedings{huang2018denselyconnectedconvolutionalnetworks,
  title={Densely Connected Convolutional Networks},
  author={Huang, Gao and Liu, Zhuang and van der Maaten, Laurens and Weinberger, Kilian Q.},
  booktitle={Proceedings of the IEEE Conference on Computer Vision and Pattern Recognition (CVPR)},
  year={2017},
  url={https://openaccess.thecvf.com/content_cvpr_2017/html/Huang_Densely_Connected_Convolutional_CVPR_2017_paper.html}
}

@inproceedings{sandler2019mobilenetv2invertedresidualslinear,
  title={MobileNetV2: Inverted Residuals and Linear Bottlenecks},
  author={Sandler, Mark and Howard, Andrew and Zhu, Menglong and Zhmoginov, Andrey and Chen, Liang-Chieh},
  booktitle={Proceedings of the IEEE Conference on Computer Vision and Pattern Recognition (CVPR)},
  year={2018},
  url={https://openaccess.thecvf.com/content_cvpr_2018/html/Sandler_MobileNetV2_Inverted_Residuals_CVPR_2018_paper.html}
}

@inproceedings{tan2020efficientnetrethinkingmodelscaling,
  title={EfficientNet: Rethinking Model Scaling for Convolutional Neural Networks},
  author={Tan, Mingxing and Le, Quoc V.},
  booktitle={International Conference on Machine Learning (ICML)},
  year={2019},
  url={https://proceedings.mlr.press/v97/tan19a.html}
}

@InProceedings{pmlr-v162-kong22a,
  title = 	 {Partial disentanglement for domain adaptation},
  author =       {Kong, Lingjing and Xie, Shaoan and Yao, Weiran and Zheng, Yujia and Chen, Guangyi and Stojanov, Petar and Akinwande, Victor and Zhang, Kun},
  booktitle = 	 {Proceedings of the 39th International Conference on Machine Learning},
  pages = 	 {11455--11472},
  year = 	 {2022},
  editor = 	 {Chaudhuri, Kamalika and Jegelka, Stefanie and Song, Le and Szepesvari, Csaba and Niu, Gang and Sabato, Sivan},
  volume = 	 {162},
  series = 	 {Proceedings of Machine Learning Research},
  month = 	 {17--23 Jul},
  publisher =    {PMLR},
  url = 	 {https://proceedings.mlr.press/v162/kong22a.html}
}

@InProceedings{Lee_2021_CVPR,
    author    = {Lee, Seunghun and Cho, Sunghyun and Im, Sunghoon},
    title     = {DRANet: Disentangling Representation and Adaptation Networks for Unsupervised Cross-Domain Adaptation},
    booktitle = {Proceedings of the IEEE/CVF Conference on Computer Vision and Pattern Recognition (CVPR)},
    month     = {June},
    year      = {2021},
    pages     = {15252-15261}
}

@ARTICLE{9437964,
  author={Deng, Wanxia and Zhao, Lingjun and Liao, Qing and Guo, Deke and Kuang, Gangyao and Hu, Dewen and Pietikäinen, Matti and Liu, Li},
  journal={IEEE Transactions on Multimedia}, 
  title={Informative Feature Disentanglement for Unsupervised Domain Adaptation}, 
  year={2022},
  volume={24},
  number={},
  pages={2407-2421},
  doi={10.1109/TMM.2021.3080516}}

@ARTICLE{10024368,
  author={Zhou, Qianyu and Gu, Qiqi and Pang, Jiangmiao and Lu, Xuequan and Ma, Lizhuang},
  journal={IEEE Transactions on Pattern Analysis and Machine Intelligence}, 
  title={Self-Adversarial Disentangling for Specific Domain Adaptation}, 
  year={2023},
  volume={45},
  number={7},
  pages={8954-8968},
  doi={10.1109/TPAMI.2023.3238727}}

@misc{levy2020largescalemethodsdistributionallyrobust,
      title={Large-Scale Methods for Distributionally Robust Optimization}, 
      author={Daniel Levy and Yair Carmon and John C. Duchi and Aaron Sidford},
      year={2020},
      eprint={2010.05893},
      archivePrefix={arXiv},
      primaryClass={math.OC},
      url={https://arxiv.org/abs/2010.05893}, 
}

@misc{li2023whacamoledilemmashortcutscome,
      title={A Whac-A-Mole Dilemma: Shortcuts Come in Multiples Where Mitigating One Amplifies Others}, 
      author={Zhiheng Li and Ivan Evtimov and Albert Gordo and Caner Hazirbas and Tal Hassner and Cristian Canton Ferrer and Chenliang Xu and Mark Ibrahim},
      year={2023},
      eprint={2212.04825},
      archivePrefix={arXiv},
      primaryClass={cs.CV},
      url={https://arxiv.org/abs/2212.04825}, 
}

@misc{arjovsky2020invariant,
      title={Invariant Risk Minimization}, 
      author={Martin Arjovsky and Léon Bottou and Ishaan Gulrajani and David Lopez-Paz},
      year={2020},
      eprint={1907.02893},
      archivePrefix={arXiv},
      primaryClass={stat.ML}
}

@misc{https://doi.org/10.48550/arxiv.2010.15775,
  doi = {10.48550/ARXIV.2010.15775},
  
  url = {https://arxiv.org/abs/2010.15775},
  
  author = {Nagarajan, Vaishnavh and Andreassen, Anders and Neyshabur, Behnam},
  
  title = {Understanding the Failure Modes of Out-of-Distribution Generalization},
  
  publisher = {arXiv},
  
  year = {2020},
  
  copyright = {arXiv.org perpetual, non-exclusive license}
}

@inproceedings{NEURIPS2020_6cfe0e61,
 author = {Shah, Harshay and Tamuly, Kaustav and Raghunathan, Aditi and Jain, Prateek and Netrapalli, Praneeth},
 booktitle = {Advances in Neural Information Processing Systems},
 editor = {H. Larochelle and M. Ranzato and R. Hadsell and M.F. Balcan and H. Lin},
 pages = {9573--9585},
 publisher = {Curran Associates, Inc.},
 title = {The Pitfalls of Simplicity Bias in Neural Networks},
 url = {https://proceedings.neurips.cc/paper/2020/file/6cfe0e6127fa25df2a0ef2ae1067d915-Paper.pdf},
 volume = {33},
 year = {2020}
}

@article{DBLP:journals/corr/abs-2004-07780,
  author       = {Robert Geirhos and
                  J{\"{o}}rn{-}Henrik Jacobsen and
                  Claudio Michaelis and
                  Richard S. Zemel and
                  Wieland Brendel and
                  Matthias Bethge and
                  Felix A. Wichmann},
  title        = {Shortcut Learning in Deep Neural Networks},
  journal      = {CoRR},
  volume       = {abs/2004.07780},
  year         = {2020},
  url          = {https://arxiv.org/abs/2004.07780},
  eprinttype    = {arXiv},
  eprint       = {2004.07780},
  bibsource    = {dblp computer science bibliography, https://dblp.org}
}

@techreport{welinder2010cub200,
  author       = {Welinder, Peter and Branson, Steve and Mita, Takeshi and Wah, Catherine and Schroff, Florian and Belongie, Serge and Perona, Pietro},
  title        = {Caltech-UCSD Birds 200},
  institution  = {California Institute of Technology},
  year         = {2010},
  number       = {CNS-TR-2010-001}
}

@inproceedings{
Sagawa2020Distributionally,
title={Distributionally Robust Neural Networks},
author={Shiori Sagawa* and Pang Wei Koh* and Tatsunori B. Hashimoto and Percy Liang},
booktitle={International Conference on Learning Representations},
year={2020},
url={https://openreview.net/forum?id=ryxGuJrFvS}
}

@inproceedings{
https://doi.org/10.48550/arxiv.2204.02937,
title={Last Layer Re-Training is Sufficient for Robustness to Spurious Correlations},
author={Polina Kirichenko and Pavel Izmailov and Andrew Gordon Wilson},
booktitle={The Eleventh International Conference on Learning Representations },
year={2023},
url={https://openreview.net/forum?id=Zb6c8A-Fghk}
}

@inproceedings{lin2014nin,
  title     = {Network In Network},
  author    = {Lin, Min and Chen, Qiang and Yan, Shuicheng},
  booktitle = {International Conference on Learning Representations (ICLR)},
  year      = {2014},
  url       = {https://openreview.net/forum?id=ylE6yojDR5yqX}
}

@inproceedings{szegedy2015inception,
  title     = {Going Deeper with Convolutions},
  author    = {Szegedy, Christian and Liu, Wei and Jia, Yangqing and Sermanet, Pierre and Reed, Scott and Anguelov, Dragomir and Erhan, Dumitru and Vanhoucke, Vincent and Rabinovich, Andrew},
  booktitle = {Proceedings of the IEEE Conference on Computer Vision and Pattern Recognition (CVPR)},
  year      = {2015}
}

\appendix
\section{Datasets}
\paragraph{Dataset Description:}
The methods are evaluated on six spurious-correlation image datasets: MNIST--CIFAR Dominoes~\cite{NEURIPS2020_6cfe0e61}, Waterbirds~\cite{welinder2010cub200, Sagawa2020Distributionally}, Spuco-Animals~\cite{joshi2023mitigating}, Spawrious~\cite{lynch2023spawrious}, CelebA~\cite{https://doi.org/10.48550/arxiv.1411.7766, Sagawa2020Distributionally}, and UrbanCars~\cite{li2023whacamoledilemmashortcutscome}. These benchmarks were selected for their widespread use in evaluating robustness to spurious correlations. Briefly:
\begin{itemize}
    \item \textbf{Dominoes} is a synthetic dataset constructed by concatenating MNIST digits (spurious) with CIFAR-10 images (core), where the label corresponds to the CIFAR-10 object class. Unlike the binary setting in the original paper, we adopt the full 10-class classification task.
    \item \textbf{Waterbirds} is a binary bird-classification dataset where the foreground bird provides the core signal, while the background environment (water or land) is spuriously correlated with the label, creating a canonical background-bias setting.
    \item \textbf{Spuco-Animals} extends the Waterbirds setting to a more challenging 4-class classification task, where the task of identifying the animal type is spuriously correlated with the background.
    \item \textbf{Spawrious} contains photorealistic AI-generated dog images, where the task of identifying the dog breed is spuriously correlated with the background style.
    \item \textbf{CelebA} contains face images where the task is to classify hair color (e.g., blond vs.\ non-blond), which is spuriously correlated with gender. Unlike background-bias datasets, the core and spurious attributes partially overlap spatially within the face region.
    \item \textbf{UrbanCars} is a car-classification benchmark in which the label is associated with multiple spurious cues, such as the background and co-occurring objects, making it a many-to-one spurious-correlation setting.
\end{itemize}

\paragraph{Terminology:}
We follow the standard terminology used in the group-robustness literature~\cite{Sagawa2020Distributionally, https://doi.org/10.48550/arxiv.2204.02937}:
\begin{itemize}
    \item \textbf{Group:} A group defined by the tuple $(y, a)$, where $y$ denotes the target label and $a$ denotes the spurious attribute.
    \item \textbf{Majority Group:} The subgroups that comprise the largest fraction of the dataset, resulting in the spurious correlation between the attribute and the label. Consequently, the spurious correlation holds for examples in the majority group.
    \item \textbf{Minority Group:} The subgroups that comprise the smallest fraction of the dataset, with attributes and class labels that conflict with the spurious correlation. 
    \item \textbf{Spurious Correlation Strength:} The probability that the spurious attribute $a$ is predictive of the target label $y$. Equivalently, this corresponds to the accuracy achieved by a classifier that relies solely on the spurious attribute.
    \item \textbf{Minority Group Accuracy:} The average test accuracy computed over samples in the minority groups, where the spurious correlation does not hold.
    \item \textbf{Majority Group Accuracy:} The average test accuracy computed over samples in the majority groups. This metric can be highly misleading, as a model can perform well on these samples by exploiting spurious correlations alone.
\end{itemize}

\paragraph{Dataset Splits:}
Each dataset is partitioned into four splits:
\begin{itemize}
    \item \textbf{Training split:} Used to learn the model parameters. The training split sizes for Dominoes, Waterbirds, CelebA, Spuco-Animals, Spawrious, and UrbanCars are 43{,}960, 9{,}472, 18{,}946, 18{,}944, 24{,}544, and 6{,}992, respectively.
    \item \textbf{Target split:} Used by DFR and DAR for post-hoc adaptation. For methods that do not use a separate target set, this split is merged into the training split. The target split contains 1{,}000 samples for all datasets.
    \item \textbf{Validation split:} Used for hyperparameter tuning. The validation split contains 1{,}000 samples for all datasets.
    \item \textbf{Test split:} Used for final evaluation.
\end{itemize}

Similar to \cite{https://doi.org/10.48550/arxiv.2110.14503}, the breakdown of the training split group counts for the spurious correlation train datasets is presented in Table \ref{Table:train_group_counts}.

\begin{table*}[tb]
\caption{Training-split group counts for the spurious-correlation datasets. For datasets with a one-to-one label--attribute correspondence, the majority groups lie on the diagonal and the minority groups lie off-diagonal. UrbanCars is unique because it is a many-to-one setting with multiple spurious cues predictive of the label.}
\label{Table:train_group_counts}
\centering
\begin{tabular}{c c c c c c c}
\toprule
\textbf{Dataset} & \textbf{Label} & \multicolumn{5}{c}{\textbf{Train Group Counts}} \\
\midrule

& \(\downarrow\) \textit{y} \(\rightarrow\) \textit{a} & Digit 0 & Digit 1 & \(\cdots\) & Digit 8 & Digit 9 \\
\cmidrule(lr){3-7}
\multirow{5}{*}{\textbf{MNIST--CIFAR Dominoes}}
& Airplane   & 4180 & 24   & \(\cdots\) & 24   & 24 \\
& Automobile & 24   & 4180 & \(\cdots\) & 24   & 24 \\
& \(\vdots\) & \(\vdots\) & \(\vdots\) & \(\ddots\) & \(\vdots\) & \(\vdots\) \\
& Ship       & 24   & 24   & \(\cdots\) & 4180 & 24 \\
& Truck      & 24   & 24   & \(\cdots\) & 24   & 4180 \\
\midrule

& \(\downarrow\) \textit{y} \(\rightarrow\) \textit{a} & Land & Water & & & \\
\cmidrule(lr){3-4}
\multirow{2}{*}{\textbf{Waterbirds}}
& \(Landbird\) & 4500 & 236 & & & \\
& \(Waterbird\) & 236  & 4500 & & & \\
\midrule

& \(\downarrow\) \textit{y} \(\rightarrow\) \textit{a} & \(Female\) & \(Male\) & & & \\
\cmidrule(lr){3-4}
\multirow{2}{*}{\textbf{CelebA}}
& \(Blonde\) & 9000 & 473 & & & \\
& \(Non-Blonde\) & 473  & 9000 & & & \\
\midrule

& \(\downarrow\) \textit{y} \(\rightarrow\) \textit{a} & Land & Water & Indoors & Outdoors & \\
\cmidrule(lr){3-6}
\multirow{4}{*}{\textbf{Spuco-Animals}}
& Landbirds  & 4500 & 236  & 0    & 0    & \\
& Waterbirds & 236  & 4500 & 0    & 0    & \\
& Small Dogs & 0    & 0    & 4500 & 236  & \\
& Big Dogs   & 0    & 0    & 236  & 4500 & \\
\midrule

& \(\downarrow\) \textit{y} \(\rightarrow\) \textit{p} & \(Jungle\) & \(Desert\) & \(Mountain\) & \(Snow\) & \\
\cmidrule(lr){3-6}
\multirow{4}{*}{\textbf{Spawrious}}
& \(Bulldog\) & 5830 & 0    & 306  & 0    & \\
& \(Corgi\) & 306  & 5830 & 0    & 0    & \\
& \(Dachshund\) & 0    & 0    & 5830 & 306  & \\
& \(Labrador\) & 0    & 306  & 0    & 5830 & \\
\midrule

& \(\downarrow\) \textit{y} \(\rightarrow\) \textit{p} & \(p=0\) & \(p=1\) & \(p=2\) & \(p=3\) & \\
\cmidrule(lr){3-6}
\multirow{2}{*}{\textbf{UrbanCars}}
& \(Country Car\) & 3156 & 166 & 166 & 8    & \\
& \(UrbanCar\) & 8    & 166 & 165 & 3153 & \\
\bottomrule
\end{tabular}
\end{table*}

\section{Experimental Setup}

\subsection{Training Setup}
\paragraph{Data Augmentation:}The datapoints are normalized using dataset-specific statistics and augmented using Random Crop and Random Horizontal Flip. The main experiment uses a randomly initialised ResNet18 model \cite{he2015deep}. This random initialisation isolates each method’s ability to learn the feature representation, independent of any pre-trained features from datasets such as ImageNet. The models are trained for \(200\) epochs to ensure convergence. The models are optimised using the Adam optimiser \cite{KingBa15} with default beta values.

\subsection{Hyperparameter Tuning Setup}
For the main experiment, we follow the hyperparameter-tuning setup described in \cite{yang2023change}. Specifically, we use the validation dataset to select the optimal hyperparameters for each method through grid search. These selected hyperparameters are then fixed and used to rerun the experiments across three random seeds, enabling us to report the mean accuracy and its standard deviation. The three main hyperparameters we tune are: learning rate in \(\{10^{-2}, 10^{-3}, 10^{-4}\}\), weight decay in \(\{10^{-1}, 10^{-2}, 10^{-3}, 10^{-4}\}\), and batch size in \(\{32, 64, 128\}\). The reported results correspond to the epoch with the highest minority-group validation accuracy, i.e., early stopping. We do not perform additional hyperparameter tuning for the ablation experiments; instead, we use the best hyperparameters identified during the main experiments.

The hyperparameter tuning for DFR and DAR follows the setup in the original DFR paper \cite{https://doi.org/10.48550/arxiv.2204.02937}. Hyperparameter tuning is performed on the base model \(ERM_{train}\), which only utilizes the training split.\footnote{In contrast, \(ERM\) uses both the training and target splits combined.} This base model is then frozen for the retraining step. During this retraining step, both the DAR and DFR steps include an additional regularization parameter that is tuned.

\subsection{Evaluation Setup}
To evaluate the model's performance, we report both majority-group accuracy and minority-group accuracy. The majority-group accuracy is not a good evaluation of the model's performance, as the model could achieve it by learning spurious correlations rather than the target function, making it misleading. By extension, the average group accuracy is also not a good metric. Therefore, minority-group accuracy is used to benchmark the model's performance by evaluating its performance on data points where the spurious correlation does not hold. This is consistent with the spurious correlation literature. All methods are repeated with three random seeds to assess robustness, and the mean and standard deviation are reported.
\section{Supplementary Metrics and Analysis for Feature Entanglement}

\subsection{Core Activation Percentage (CAP) Metric}
Given an output feature map of the final convolutional layer, the activation in the top half corresponds to the MNIST input. In contrast, the activation at the bottom half of the feature map corresponds to the CIFAR input. We compute the \textit{Core Activation Percentage (CAP)} for the \(j\)-th feature map as follows:
\begin{equation}
    \label{eqn:cap}
    CAP_j = \mathbb{E}_{i}\left[ \frac{\sum_{h=H/2}^{H}\sum_{w}^{W}|\mathbf{A}_i[j, h, w]|}{\sum_{h=1}^{H}\sum_{w=1}^{W}|\mathbf{A}_i[j, h, w]|}\right] * 100\%
\end{equation}
where \(H\) and \(W\) are the height and width of the feature map, and \(\mathbf{A}_i \in \mathbb{R}^{d \times H \times W}\) is the activation of the feature map for input \(\mathbf{x}_i\). The CAP metric is computed for all of the model's output feature maps.

\subsection{Core GradCAM Percentage (CGP) Metric}
We use GradCAM \cite{Selvaraju_2019} to obtain a heatmap that highlights the regions of the input image the model focuses on when making its prediction. As with the CAP, the top half of the GradCAM heatmap corresponds to the MNIST input, whereas the bottom half corresponds to the CIFAR input. We measure the \textit{Core GradCAM Percentage (CGP)} as follows:
\begin{equation}
    \label{eqn:cgp}
    CGP = \mathbb{E}_{i}\left[ \frac{\sum_{h=H/2}^{H}\sum_{w=1}^{W}|\mathbf{G}_i[h, w]|}{\sum_{h=1}^{H}\sum_{w=1}^{W}|\mathbf{G}_i[h, w]|}\right] * 100\%
\end{equation}
where \(H\) and \(W\) are the height and width of the GradCAM heatmap, and \(\mathbf{G}_i \in \mathbb{R}^{H \times W}\) is the GradCAM heatmap for input \(\mathbf{x}_i\). The CGP metric is computed for the model's output.

\subsection{Ablation Results and Analysis}
\begin{table*}[tb]
    \centering
    \resizebox{\textwidth}{!}{
    \begin{tabular}{c c c c c c c c c c}
        \toprule
        \textbf{Method} & \(\mathbf{ERM_{Core}}\) & \(\mathbf{ERM_{train}}\) & \(\mathbf{DFR}\) & \(\mathbf{DFR_{FC}}\) & \(\mathbf{DFR_{CNN}}\) & \(\mathbf{DAR}\) &\(\mathbf{DAR_{Spu}}\) \\
        \toprule
        \(\mathbf{CGP}\) &\(91.0 \pm 0.3\) & \(66.8 \pm 0.4\) & \(70.0 \pm 1.5\) & \(82.5 \pm 1.2\) &\(79.6 \pm 1.8\) &\(95.8 \pm 0.5\) & \(2.2 \pm 0.5\)\\
        \bottomrule
    \end{tabular}
    }
    \caption{Core GradCAM Percentage for Model's Output Prediction.}
    \label{Table:CGP_ablation}
\end{table*}

\begin{figure}[tb]
  \begin{subfigure}{0.30\textwidth}
    \includegraphics[width=\linewidth]{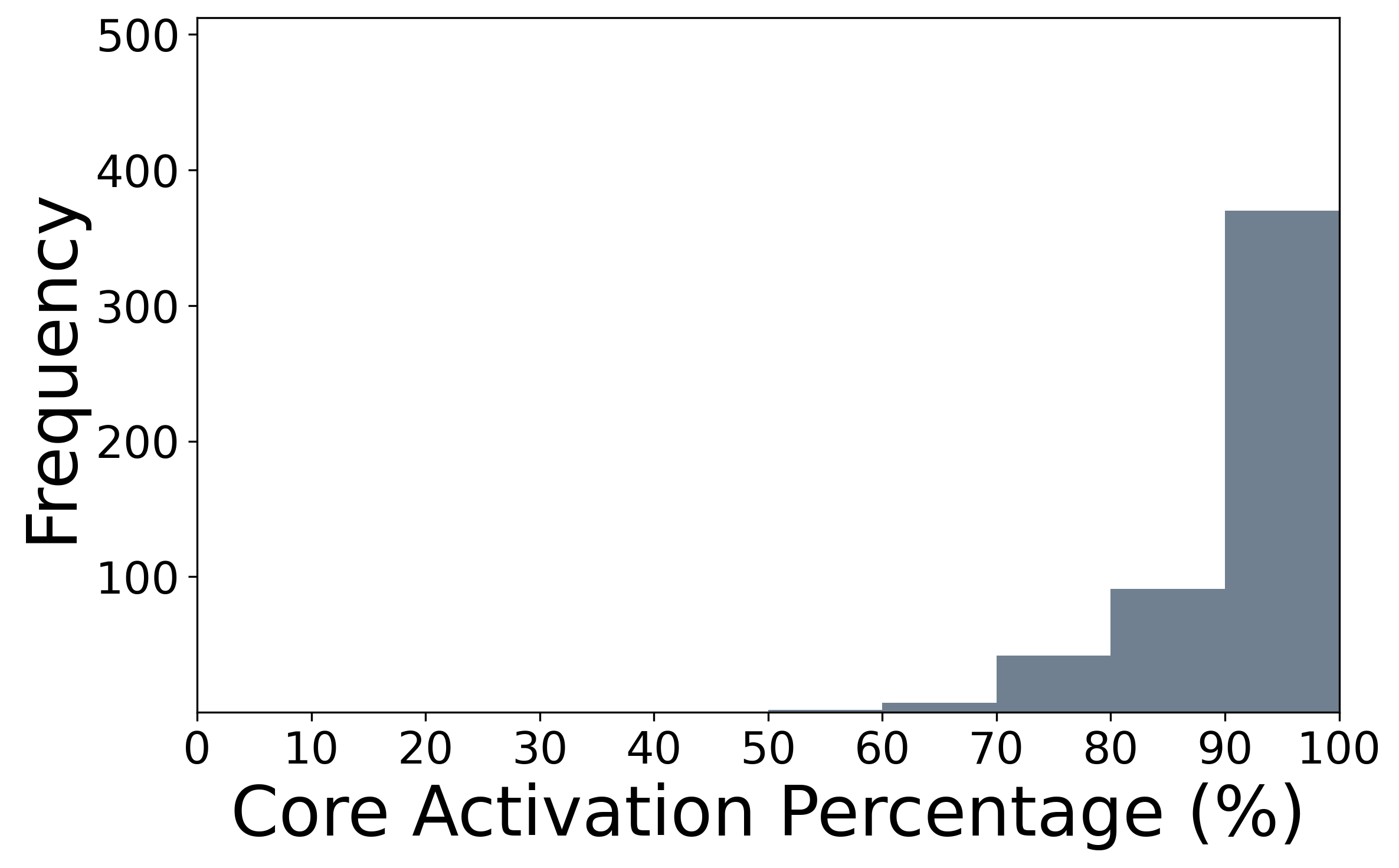}\hfill
    \caption{\(ERM_{Core}\)}
      \label{fig:map_cap_ERM_core}
  \end{subfigure}%
  \hspace*{\fill}   
  \begin{subfigure}{0.30\textwidth}
  \includegraphics[width=\linewidth]{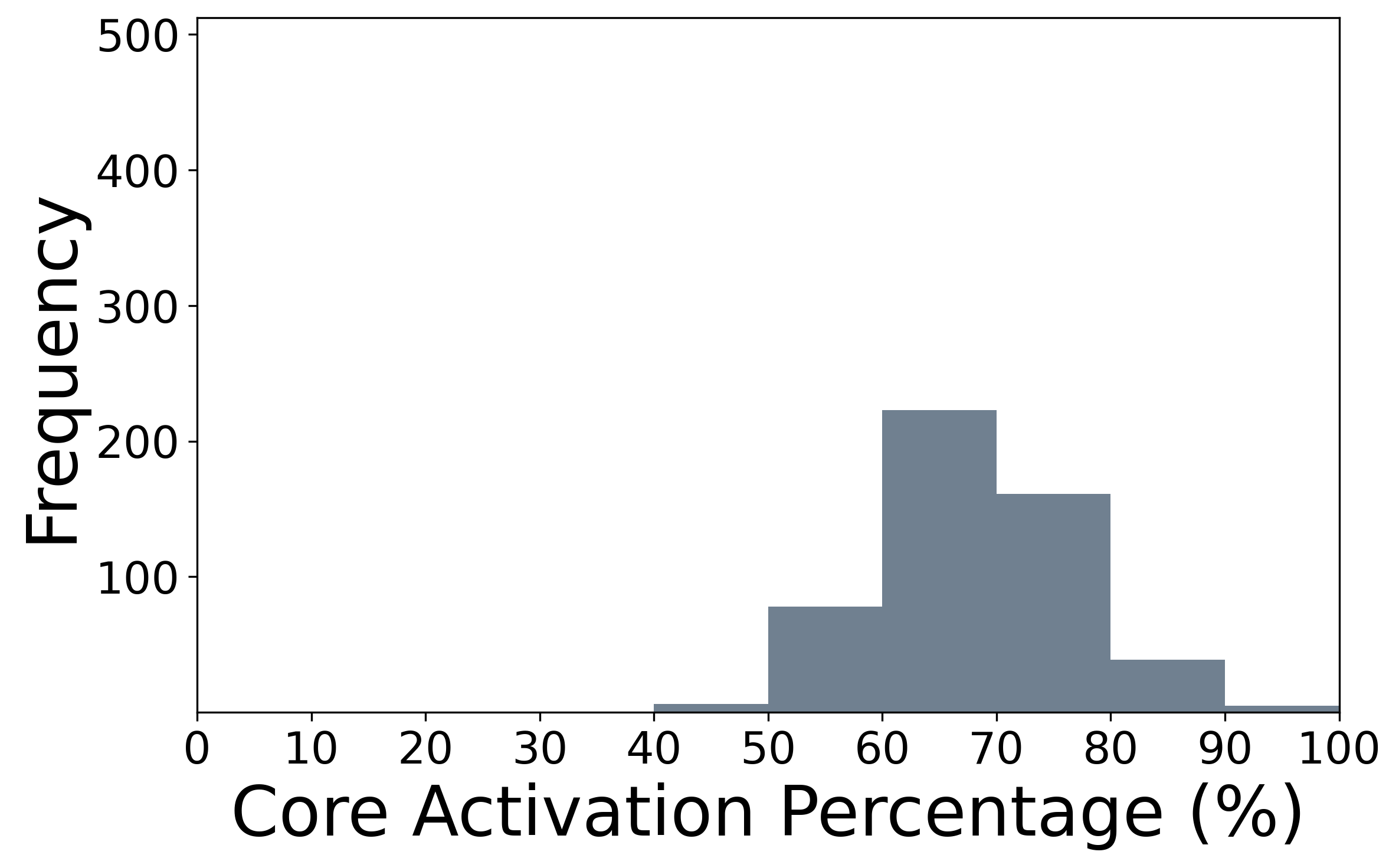}\hfill
  \caption{\(ERM\)}
  \label{fig:map_cap_ERM}
  \end{subfigure}\hspace*{\fill}   
  \begin{subfigure}{0.30\textwidth}
  \includegraphics[width=\linewidth]{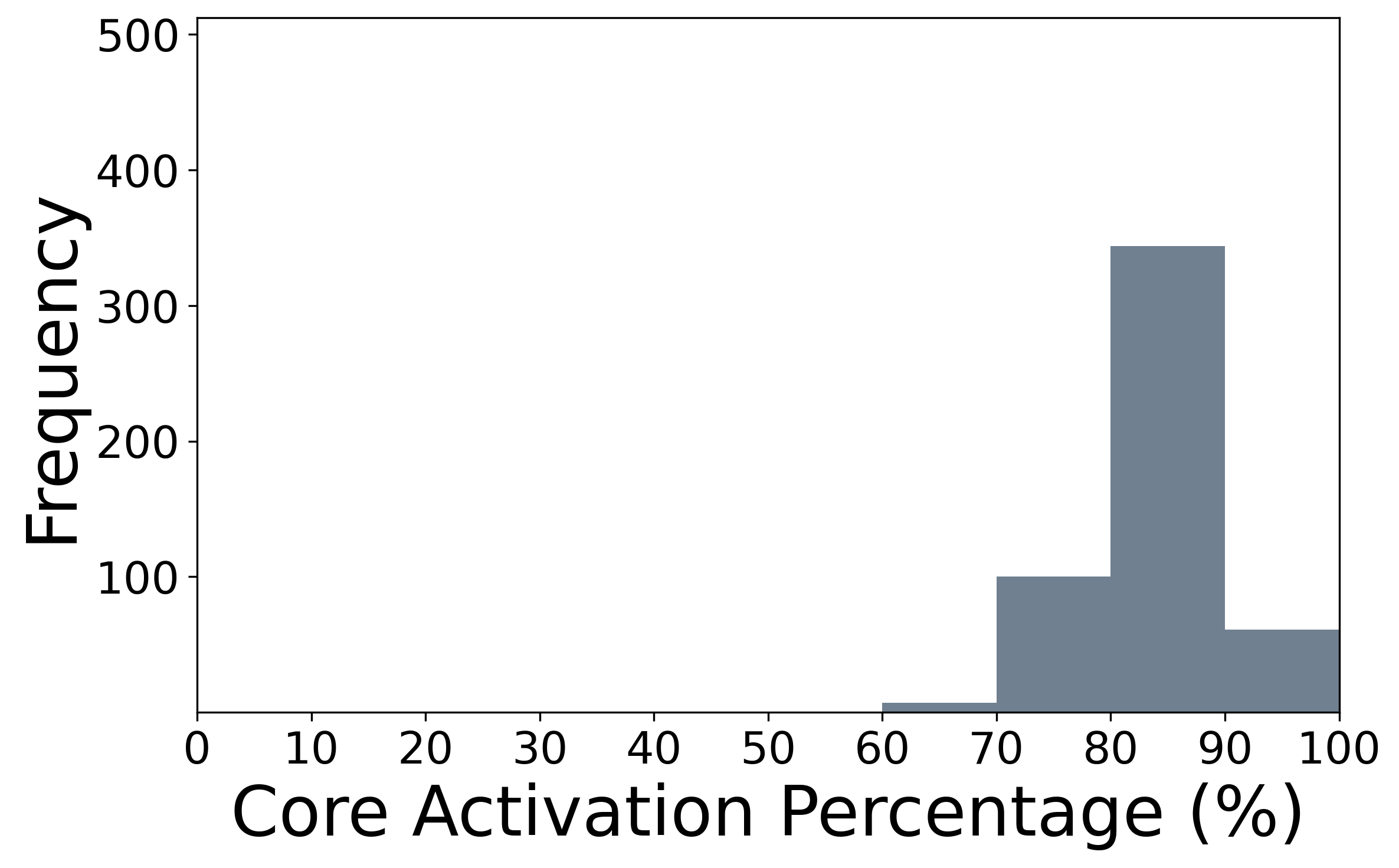}\hfill
  \caption{\(DFR_{CNN}\)} 
  \label{fig:map_cap_cnn}
  \end{subfigure}
\caption{Histogram of CAP values across 512 output feature maps.} \label{fig:map_cap}
\end{figure}
The results from these two additional ablation metrics (Table \ref{Table:CGP_ablation} and Figure \ref{fig:map_cap}) support the conclusions drawn in Section 4:
\begin{enumerate}
    \item \(ERM_{core}\): When trained on a dataset where the spurious correlation is absent, the feature maps (Figure \ref{fig:map_cap_ERM_core}) and output prediction ($CGP=91.0\%$, Table \ref{Table:CGP_ablation}) demonstrate a strong reliance on core features.
    \item \(ERM\): When trained on the dataset with spurious correlation, the feature maps (Figure \ref{fig:map_cap_ERM}) exhibit significant entanglement between core and spurious features, resulting in an output that depends on both ($CGP=66.8\%$, Table \ref{Table:CGP_ablation}).
    \item \(DFR\): While DFR improves the CGP score ($CGP=70.0\%$, Table \ref{Table:CGP_ablation}) compared to $ERM_{train}$, it is still insufficient to fully eliminate the influence of spurious features as it still relies on the entangled feature maps (Figure \ref{fig:map_cap_ERM}).
    \item \(DFR_{CNN}\): Modifying $DFR$ to include the retraining of the final CNN layer reduces reliance on the spurious features in both the feature maps (Figure \ref{fig:map_cap_cnn}) and the output prediction ($CGP=79.6\%$, Table \ref{Table:CGP_ablation}), but remains insufficient to fully remove their influence.
    \item \(DFR_{FC}\): Modifying $DFR$ to retrain a two-layer classification head (instead of the original single-layer) reduces the reliance on the spurious feature in the output prediction ($CGP=82.5\%$, Table \ref{Table:CGP_ablation}), but remains insufficient to fully remove their influence.
    \item \(DAR\): The proposed $DAR$ method significantly outperforms $DFR$, effectively extracting core features while almost completely eliminating spurious feature influence ($CGP=95.8\%$, Table \ref{Table:CGP_ablation}). Alternatively, it can be configured to extract only the spurious features ($CGP=2.2\%$, Table \ref{Table:CGP_ablation}).
\end{enumerate}

\section{GradCAM Images}
\begin{figure}[tb]
    \centering
  \begin{subfigure}{0.23\textwidth}
    \includegraphics[width=0.24\linewidth]{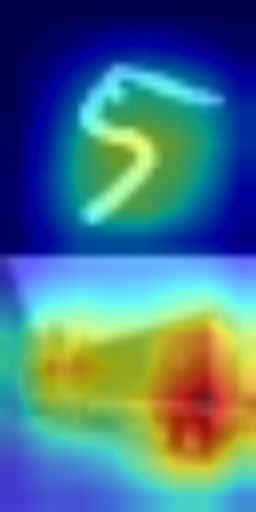}\hfill
    \includegraphics[width=0.24\linewidth]{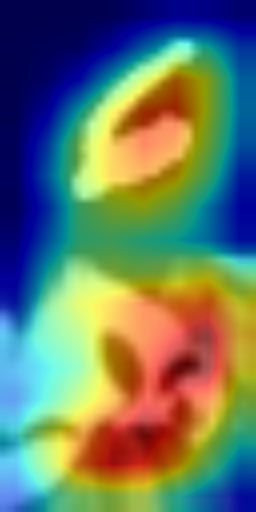}\hfill
    \includegraphics[width=0.24\linewidth]{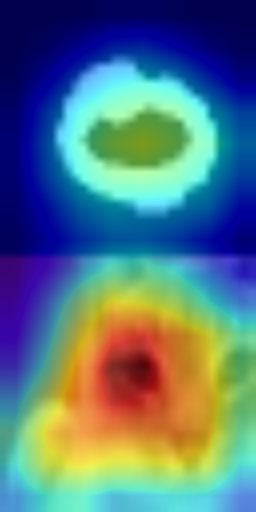}\hfill
    \includegraphics[width=0.24\linewidth]{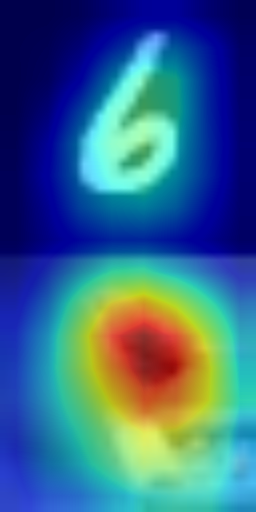}\\
    \includegraphics[width=0.24\linewidth]{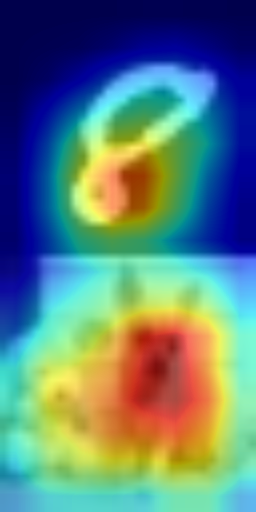}\hfill
    \includegraphics[width=0.24\linewidth]{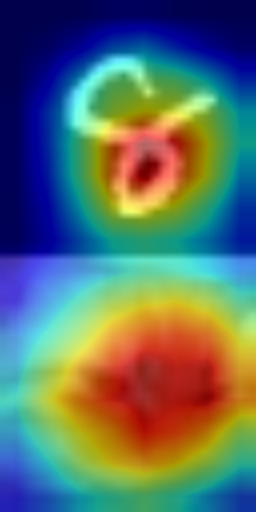}\hfill
    \includegraphics[width=0.24\linewidth]{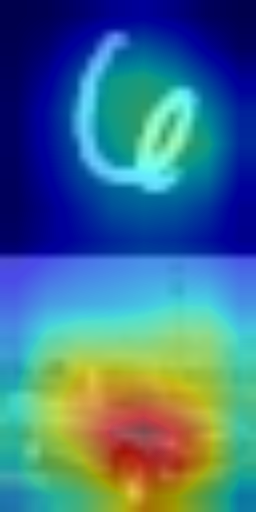}\hfill
    \includegraphics[width=0.24\linewidth]{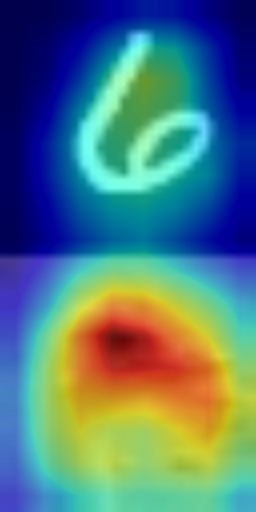}\\
    \includegraphics[width=0.24\linewidth]{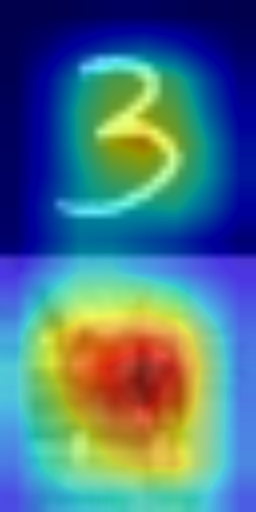}\hfill
    \includegraphics[width=0.24\linewidth]{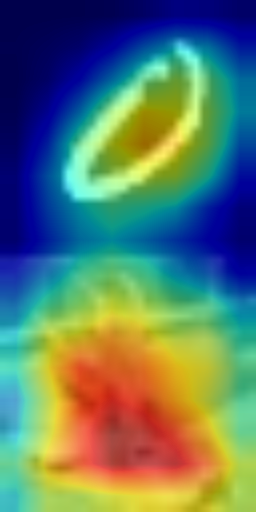}\hfill
    \includegraphics[width=0.24\linewidth]{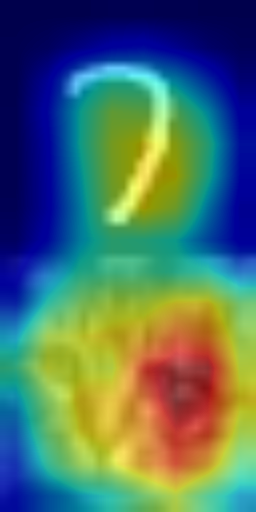}\hfill
    \includegraphics[width=0.24\linewidth]{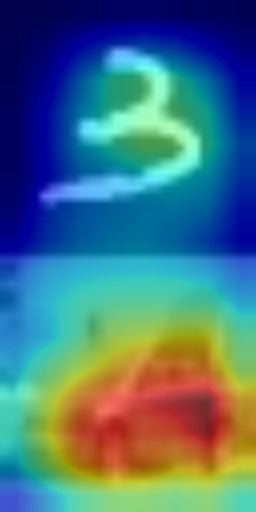}\\
    \includegraphics[width=0.24\linewidth]{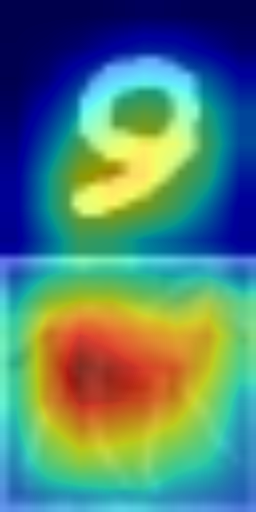}\hfill
    \includegraphics[width=0.24\linewidth]{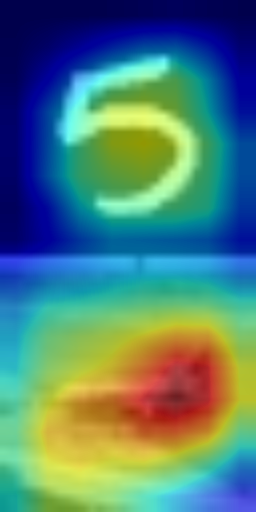}\hfill
    \includegraphics[width=0.24\linewidth]{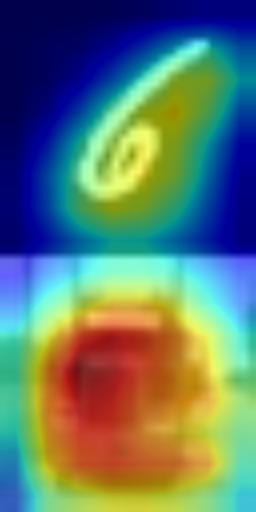}\hfill
    \includegraphics[width=0.24\linewidth]{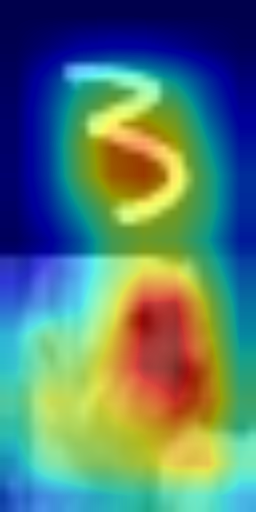}
    \caption*{1(a) $ERM$} 
  \end{subfigure}%
  \hspace*{\fill}   
  \begin{subfigure}{0.23\textwidth}
    \includegraphics[width=0.24\linewidth]{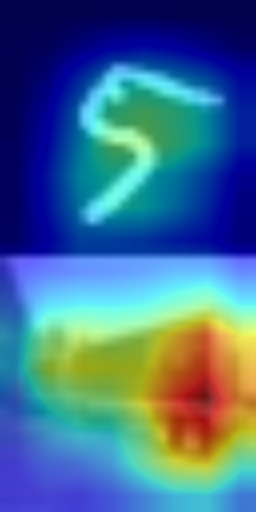}\hfill
    \includegraphics[width=0.24\linewidth]{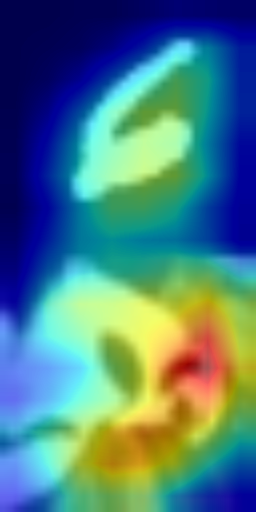}\hfill
    \includegraphics[width=0.24\linewidth]{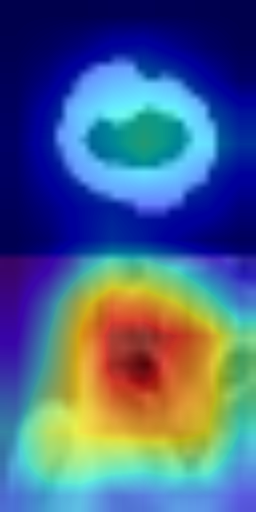}\hfill
    \includegraphics[width=0.24\linewidth]{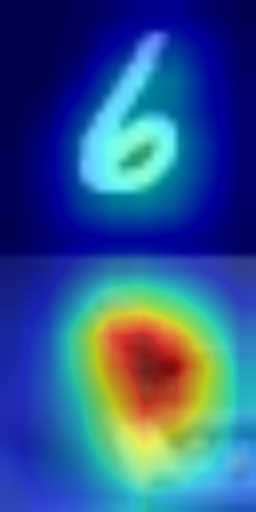}\\
    \includegraphics[width=0.24\linewidth]{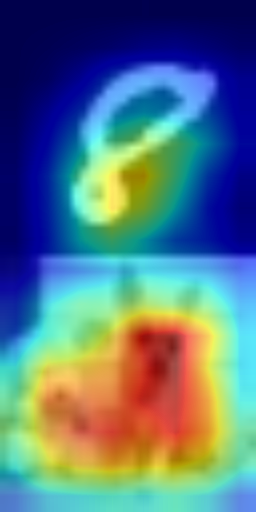}\hfill
    \includegraphics[width=0.24\linewidth]{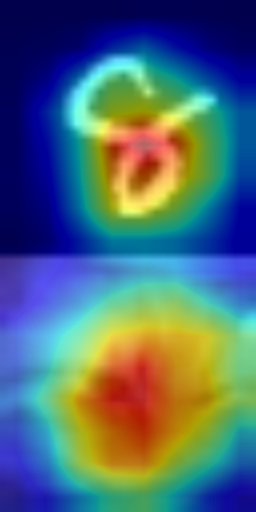}\hfill
    \includegraphics[width=0.24\linewidth]{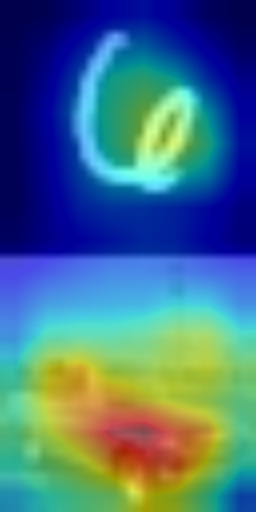}\hfill
    \includegraphics[width=0.24\linewidth]{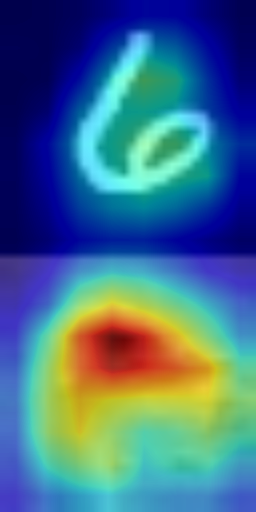}\\
    \includegraphics[width=0.24\linewidth]{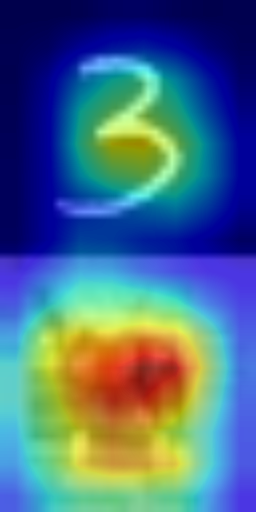}\hfill
    \includegraphics[width=0.24\linewidth]{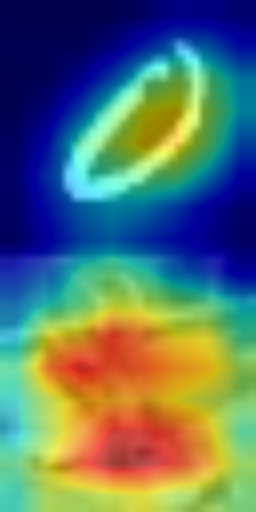}\hfill
    \includegraphics[width=0.24\linewidth]{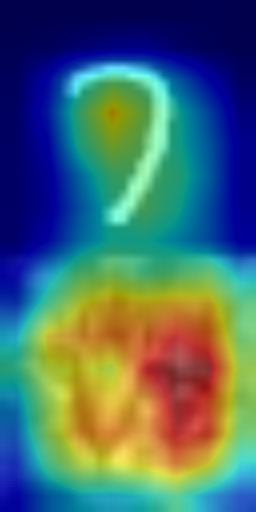}\hfill
    \includegraphics[width=0.24\linewidth]{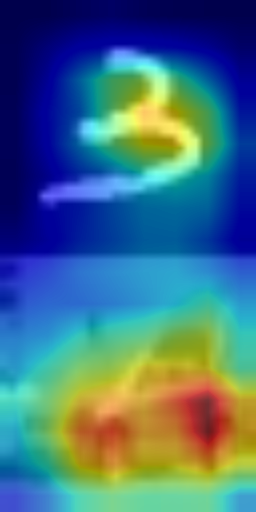}\\
    \includegraphics[width=0.24\linewidth]{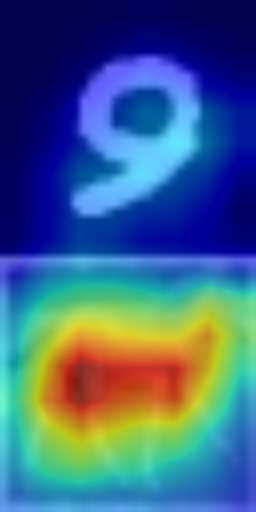}\hfill
    \includegraphics[width=0.24\linewidth]{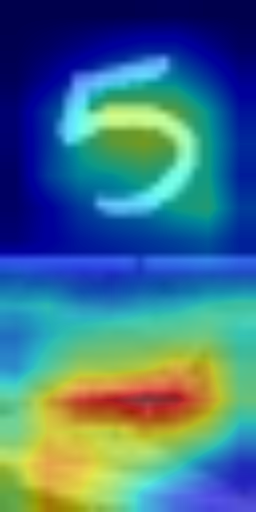}\hfill
    \includegraphics[width=0.24\linewidth]{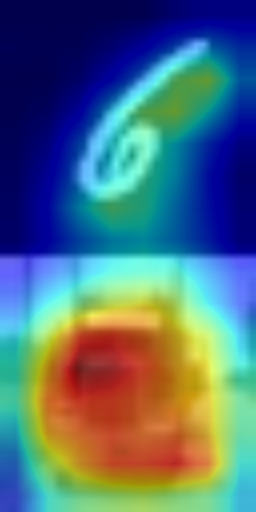}\hfill
    \includegraphics[width=0.24\linewidth]{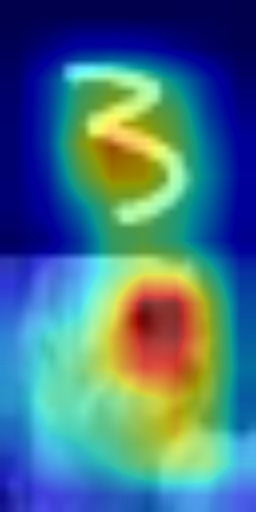}
    \caption*{1(b) $DFR$} 
  \end{subfigure}%
\hspace*{\fill}   
\begin{subfigure}{0.23\textwidth}
    \includegraphics[width=0.24\linewidth]{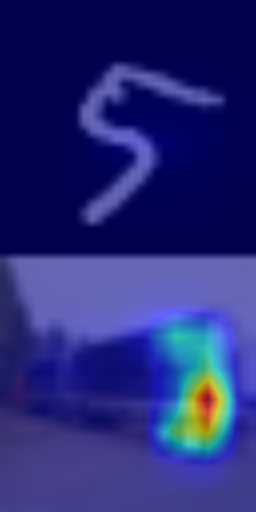}\hfill
    \includegraphics[width=0.24\linewidth]{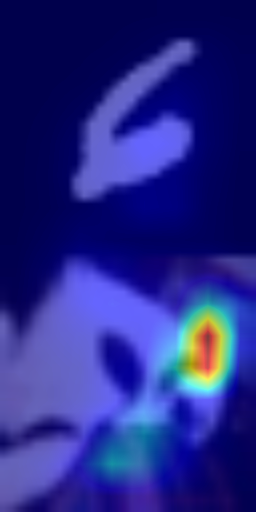}\hfill
    \includegraphics[width=0.24\linewidth]{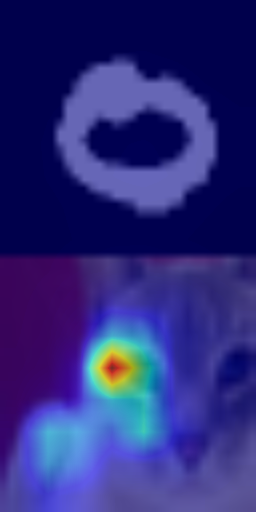}\hfill
    \includegraphics[width=0.24\linewidth]{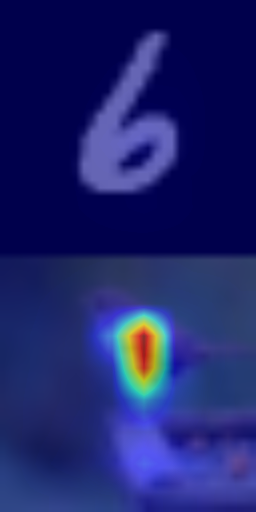}\\
    \includegraphics[width=0.24\linewidth]{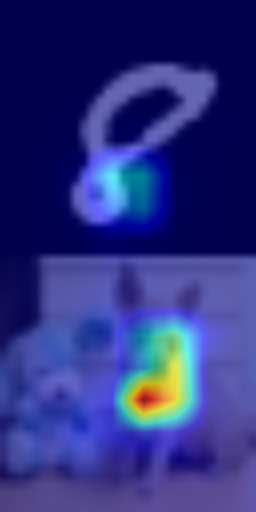}\hfill
    \includegraphics[width=0.24\linewidth]{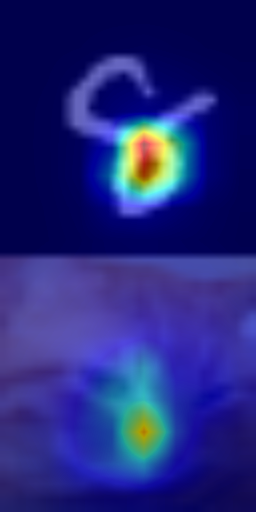}\hfill
    \includegraphics[width=0.24\linewidth]{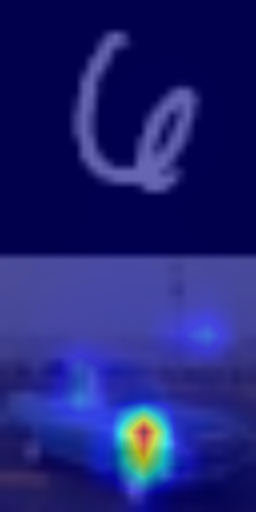}\hfill
    \includegraphics[width=0.24\linewidth]{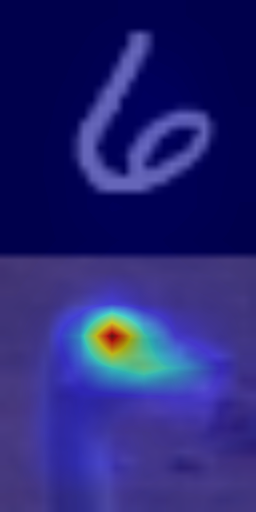}\\
    \includegraphics[width=0.24\linewidth]{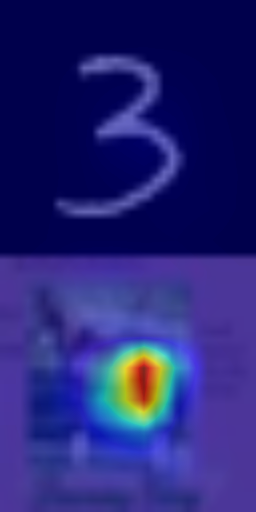}\hfill
    \includegraphics[width=0.24\linewidth]{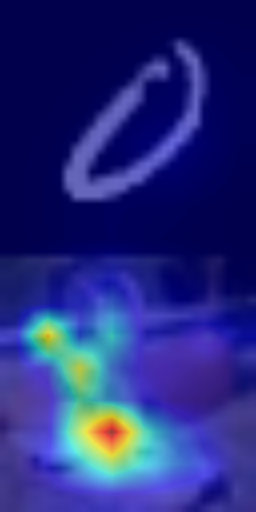}\hfill
    \includegraphics[width=0.24\linewidth]{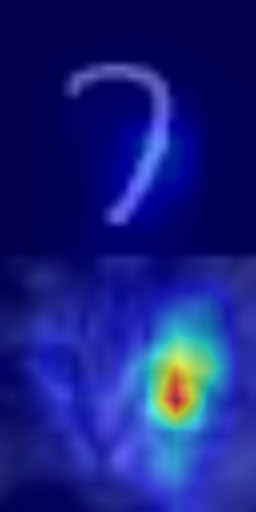}\hfill
    \includegraphics[width=0.24\linewidth]{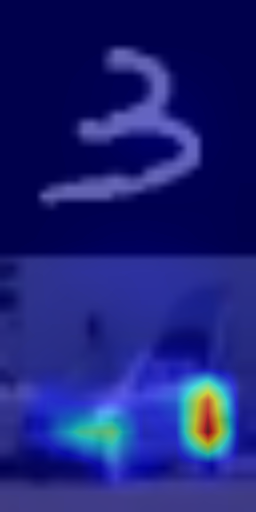}\\
    \includegraphics[width=0.24\linewidth]{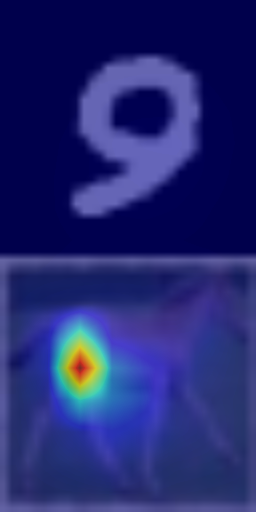}\hfill
    \includegraphics[width=0.24\linewidth]{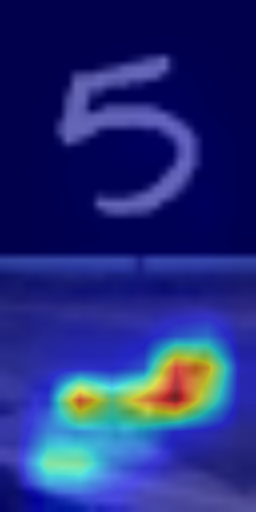}\hfill
    \includegraphics[width=0.24\linewidth]{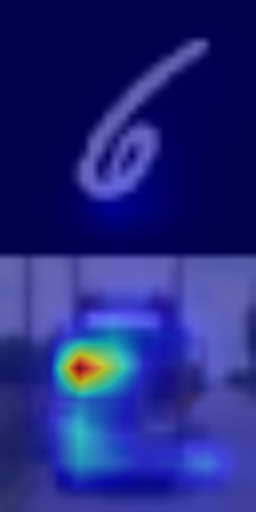}\hfill
    \includegraphics[width=0.24\linewidth]{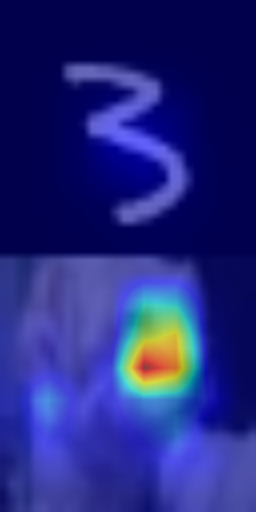}
    \caption*{1(c) $DAR$}
  \end{subfigure}%
  \hspace*{\fill}   
  \begin{subfigure}{0.23\textwidth}
    \includegraphics[width=0.24\linewidth]{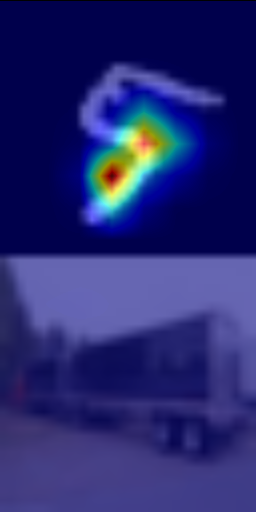}\hfill
    \includegraphics[width=0.24\linewidth]{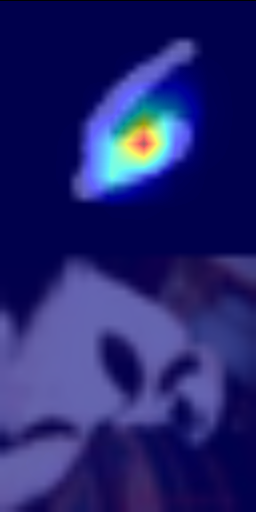}\hfill
    \includegraphics[width=0.24\linewidth]{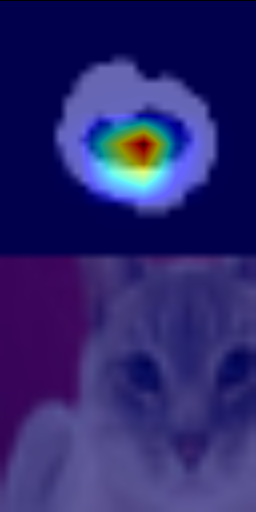}\hfill
    \includegraphics[width=0.24\linewidth]{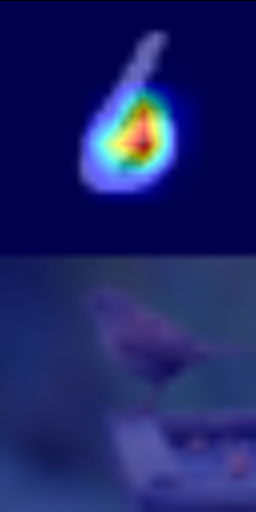}\\
    \includegraphics[width=0.24\linewidth]{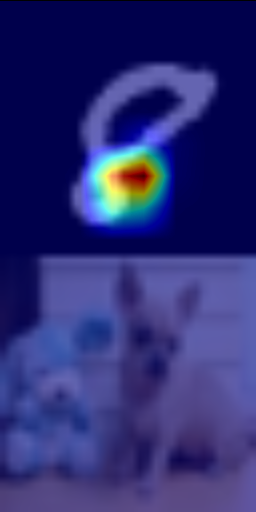}\hfill
    \includegraphics[width=0.24\linewidth]{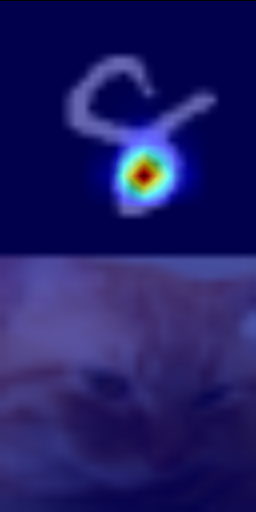}\hfill
    \includegraphics[width=0.24\linewidth]{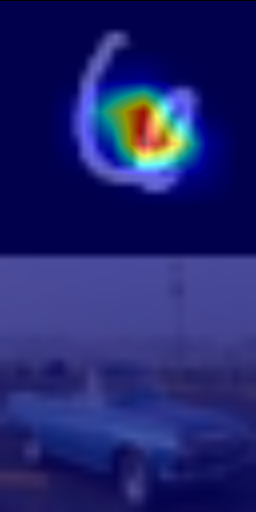}\hfill
    \includegraphics[width=0.24\linewidth]{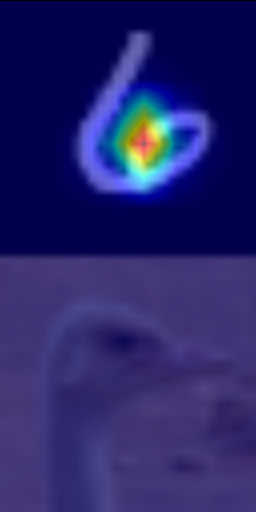}\\
    \includegraphics[width=0.24\linewidth]{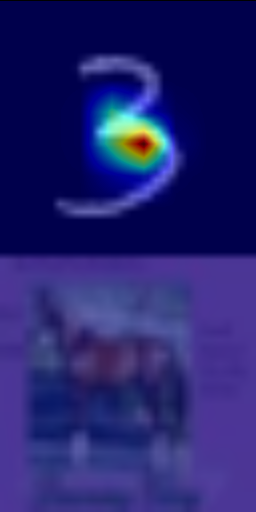}\hfill
    \includegraphics[width=0.24\linewidth]{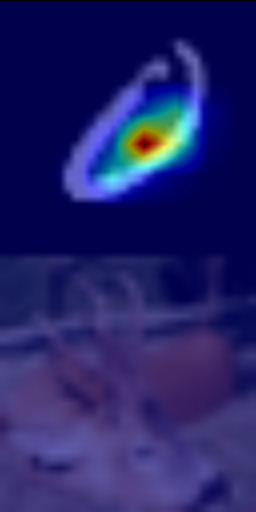}\hfill
    \includegraphics[width=0.24\linewidth]{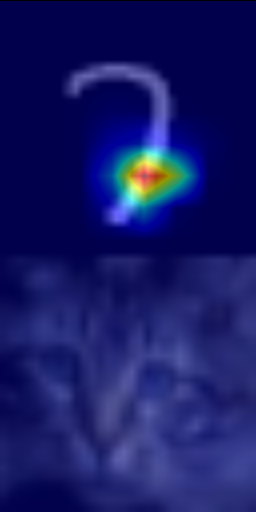}\hfill
    \includegraphics[width=0.24\linewidth]{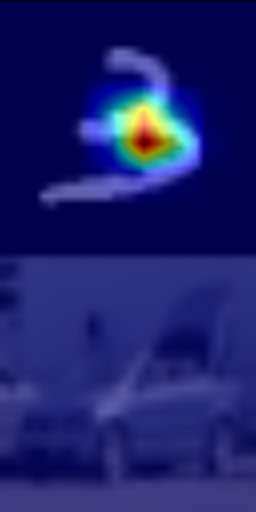}\\
    \includegraphics[width=0.24\linewidth]{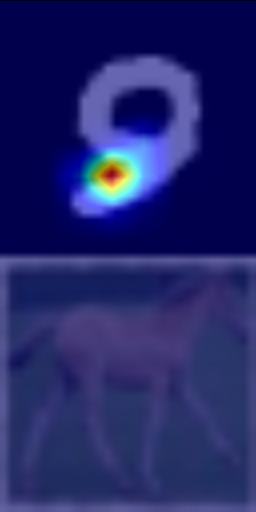}\hfill
    \includegraphics[width=0.24\linewidth]{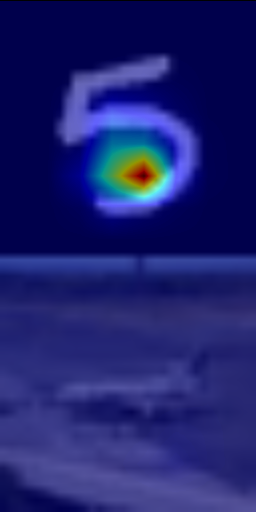}\hfill
    \includegraphics[width=0.24\linewidth]{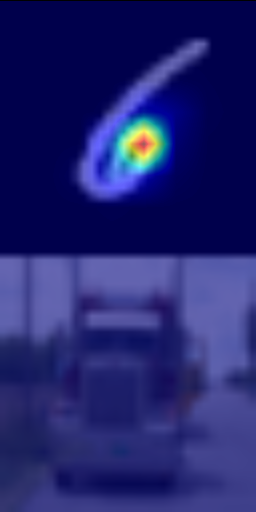}\hfill
    \includegraphics[width=0.24\linewidth]{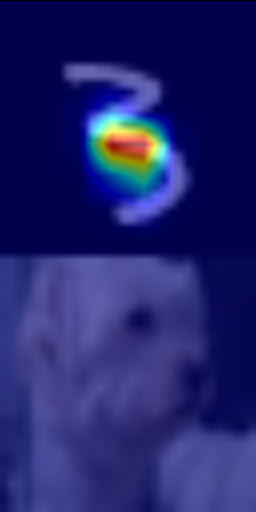}
    \caption*{1(d) $DAR_{Spu}$}
  \end{subfigure}%

\caption{A random sample of 16 GradCAM images from the test datasets for $ERM$, $DFR$, $DAR$, $DAR_{Spu}$ models that were obtained from the main experiments for the Dominoes dataset.} \label{fig:gradcam}
\end{figure}

Through visual inspection of Figure \ref{fig:gradcam}, we observe that the ERM model relies on both the core and spurious features. The DFR model reduces reliance on spurious features, but it is insufficient to completely remove their influence due to feature entanglement. Finally, our proposed DAR method can almost completely eliminate the influence of these spurious features, focusing almost exclusively on the core features. Alternatively, it can predict the spurious attributes and focus on the spurious features. We also note that the attention mechanism learns to assign dynamic, highly focused attention, with attention for each input image confined to the most discriminative regions.

\section{Mathematical Formulation of DAR}
In Section 4, we introduced the architectural design used in our proposed DAR model. In this appendix, we mathematically introduce the architecture, which is based on the scaled dot-product attention mechanism:

\begin{equation}
    SDPA(\mathbf{Q}, \mathbf{K}, \mathbf{V}) = softmax \left( \frac{\mathbf{Q}\mathbf{K}^T}{\sqrt{d}}\right)\mathbf{V}
\end{equation}
where $\mathbf{Q}, \mathbf{K}, \mathbf{V}$ represent the query, key, and value matrices, respectively, while $d$ represents the dimension of the vectors. The multihead attention architecture extends the $SDPA$ as follows:
\begin{align}
    head_t = SDPA(\mathbf{Q}\mathbf{W}_Q^t, \mathbf{KW}_K^t, \mathbf{VW}_V^t), \quad t=1,\ldots,n_{\text{head}}\\
    MHA(\mathbf{Q}, \mathbf{K}, \mathbf{V}) = Concat(head_1, ..., head_{n_{head}})\mathbf{W}_O
\end{align}
where $n_{head}$ refers to the number of heads and $\mathbf{W}_Q^t, \mathbf{W}_K^t, \mathbf{W}_V^t \in R^{d, d_{head}}$ are linear projection matrices that map from the original dimension $d$ to the dimension of each head $d_{head} = d / n_{head}$. $\mathbf{W}_O \in R^{d, d}$ is the output projection matrix.

Using these attention modules, we can define our two-layer attention architecture as:
\begin{align}
    \mathbf{Q}' &= MHA(\mathbf{Q}, \mathbf{A}, \mathbf{A})\\
    \mathbf{H} &= MHA(\mathbf{Q}', \mathbf{A}, \mathbf{A})
\end{align}
where $\mathbf{Q}\in\mathbb{R}^{k\times d}$ represents $k$ learnable query vectors, $\mathbf{A} \in \mathbb{R}^{HW\times d}$ represents the output activation map, which is used as both the key and value matrices. In this formulation, the first layer updates the query vectors in the context of the feature map, before using the updated queries to extract the core features.

Finally, $f^{attention}$ represents the feature embedding extracted from the feature map $\mathbf{A}$ based on the learnable query vectors $\mathbf{Q}$. We take the average of these feature vectors to obtain the final feature representation:
\begin{equation}
\label{eqn:average}
    h = \frac{1}{k}\sum_{i=1}^k \mathbf{H}[i] \in \mathbb{R}^d
\end{equation}


\end{document}